\documentclass[journal]{IEEEtran}
\usepackage{times}
\usepackage{epsfig}
\usepackage{graphicx}
\usepackage{amsmath}
\usepackage{amssymb}
\usepackage[symbol]{footmisc}
\usepackage{subfigure}
\usepackage[normalem]{ulem}
\usepackage[table]{xcolor}
\usepackage{multirow}
\usepackage{graphicx}
\usepackage{svg}
\usepackage{subfig}
\usepackage{array}
\usepackage{color,soul}
\usepackage{booktabs}
\usepackage{tabu}
\usepackage[noend]{algpseudocode}
\usepackage[ruled,vlined,boxed]{algorithm2e}
\usepackage[toc]{glossaries}
\usepackage{url}
\usepackage{amssymb} 
\usepackage{amsmath}
\usepackage{wrapfig}
\usepackage{mathtools}

\DeclareMathOperator{\argmin}{argmin} 
\DeclareMathOperator{\argmax}{argmax} 
\def\mathbi#1{\textbf{\em #1}}

\begin{document}

\title{Quantized Embedding Vectors for Controllable Diffusion Language Models}

\author{Cheng~Kang, Xinye~Chen, Yong~Hu,~\IEEEmembership{Senior Member,~IEEE},  Daniel~Novak,~\IEEEmembership{Senior Member,~IEEE}
\thanks{This work was supported in part by the Research Centre for Informatics (Grant No. CZ.02.1.01/0.0/0.0/16\_019/0000765) in part by the Brain Dynamics (Grant No. CZ.02.01.01/00/22\_008/0004643), in part by the Student Grant Agency of the Czech Technical University in Prague (Grant No. SGS22/165/OHK3/3T/13). The corresponding author is Cheng Kang. }
\thanks{C. Kang and D. Novak are with the Department of Cybernetics, Faculty of Electrical Engineering, Czech Technical University in Prague, Prague, Czech Republic (e-mail: kangchen@fel.cvut.cz, xnovakd1@fel.cvut.cz).}
\thanks{X. Chen is with the Department of Numerical Mathematics, Charles University, Praha, Czech Republic (e-mail:  xinye.chen@mff.cuni.cz).}
\thanks{Y. Hu is with the Department of Orthopaedics and Traumatology, University of Hong Kong, Hong Kong (e-mail: yhud@hku.hk).}
}


\maketitle

\begin{abstract}
Improving the controllability, portability, and inference speed of diffusion language models (DLMs) is a key challenge in natural language generation. While recent research has shown significant success in complex text generation with language models, the memory and computational power are still very demanding and fall short of expectations, which naturally results in low portability and instability for the models. To mitigate these issues, numerous well-established methods were proposed for neural network quantization. To further enhance their portability of independent deployment as well as improve their stability evaluated by language perplexity, we propose a  novel approach called the Quantized Embedding Controllable Diffusion Language Model (QE-CDLM). QE-CDLM builds upon the recent successful controllable DLMs by remodeling the task-specific embedding space via quantization. This leads to a gradient-based controller for the generation tasks, and more stable intermediate latent variables are obtained, which naturally brings in an accelerated convergence as well as better controllability. Additionally, the adaption fine-tuning method is employed to reduce tunable weights. Experimental results on five challenging fine-grained control tasks demonstrate that QE-CDLM compares favorably to existing methods in terms of quality and feasibility, achieving better perplexity and lightweight fine-tuning.
\end{abstract}


%
\IEEEpeerreviewmaketitle

\section{Introduction}

Large language models (LLMs) have shown their ability to generate high quality text \cite{chowdhery2022palm,zhang2022opt} and produce task-controllable outputs \cite{yang2021fudge,li2022diffusion}. Fine-tuning methods (e.g., \cite{hu2021lora,he2021towards}) can be employed to control LLMs based on supervised data, such as control text \cite{keskar2019ctrl}. This involves freezing pre-trained LLMs and generating texts using an external classifier, leading to lightweight and modular plug-and-play language models (PPLMs) \cite{pascual2021plug}. 
Nevertheless, achievements in control thus far have been restricted to basic attribute-level controls, such as sentiment or topic. \cite{yang2021fudge,pascual2021plug}, and there is a need to improve their portability, convergence speed and perplexity \cite{yang2021fudge,li2022diffusion}. The existing practical challenge refers to optimizing the embedding and accelerating the generation process. The learned embedding defines a mapping that bridges the discrete text and the continuous input, and its quality affects the language perplexity. Accordingly, in the text generation process, (1) obtaining generation language models (GLMs) from scratch should be avoided; (2) the generating process should be facilitated; and the embedding space should be optimized by processing learned embedding vectors, as pre-trained and learned embedding vectors outperform randomly initialized embedding vectors on controllable DLMs (Diffusion Language Models) \cite{li2022diffusion}; and (3) GLMs should be obtained using parameter-efficient fine-tuning (FT) methods to reduce the tunable weights.

\begin{figure}
    \begin{center}
    \includegraphics[width=0.48\textwidth]{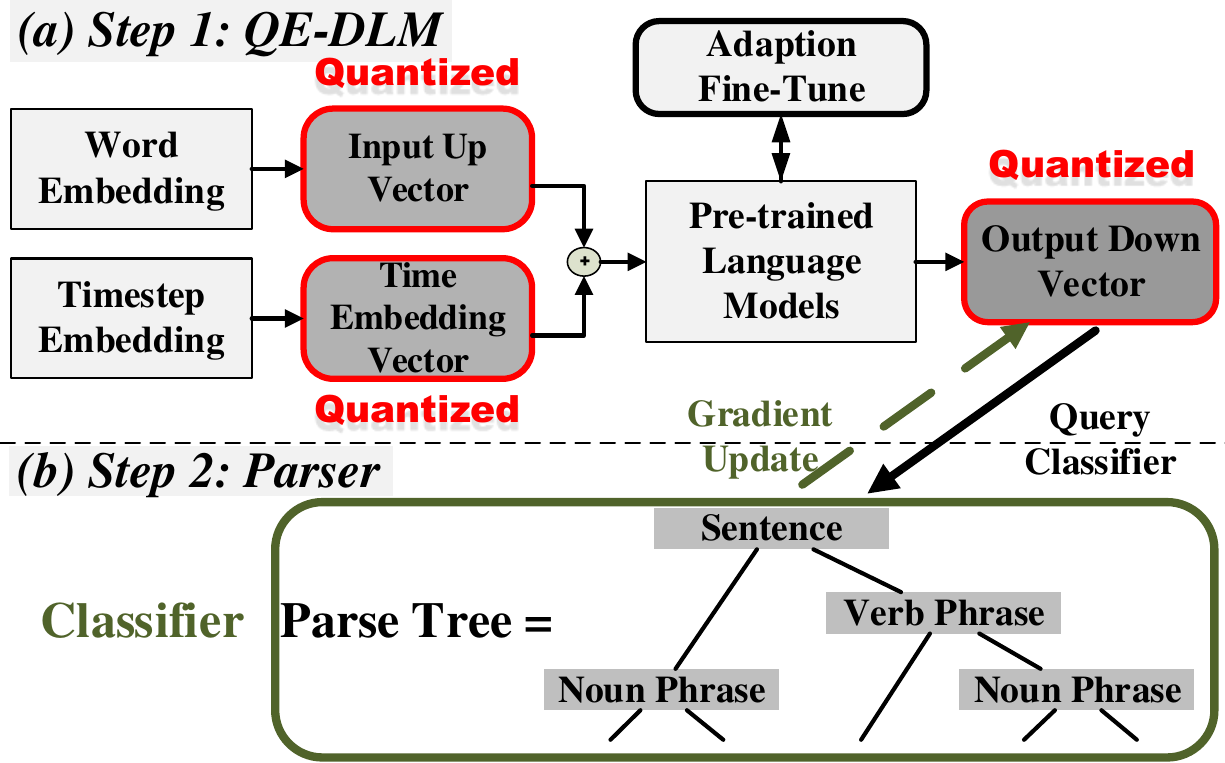}
    \end{center}
    \caption{The proposed method contains two main steps: QE-DLM and Classifier. In the first step, QE-DLM denoises a sequence of quantized Gaussian vectors that are added to word vectors. The quantized embedding vectors then compress and remodel the discrete latent space through a reverse diffusion process. In the second step, the Classifier updates the gradient on the continuous latent space using control. The DLM demonstrates its capability to generate fluent text, and the proper classifier effectively constrains the generated text based on specific control dependence, such as a Parse Tree.}
    \label{Figure3-1}
\end{figure}

Neural network quantization has revolutionized modern deep learning by alleviating the issue of pressing demand for computing resources and power due to increasing size and training time of neural networks. Quantization methods in combination with continuous diffusion language models also have shown success in various domains such as computer vision \cite{gu2022vector}, audio and music generation \cite{zhu2022discrete, yang2022diffsound}. However, applying these methods to text generation has been challenging due to the discrete nature of text \cite{li2022diffusion}. On the other hand, adapter tuning methods have been widely used to fine-tune pre-trained language models by incorporating small neural modules into them, see \cite{hu2021lora,pfeiffer2020adapterfusion,li2021prefix} and references therein. Despite the effectiveness of this fine-tuning technique, it has not been explored in the context of controllable DLMs.

\begin{figure*}[t]
\begin{center}
\includegraphics[scale=0.52]{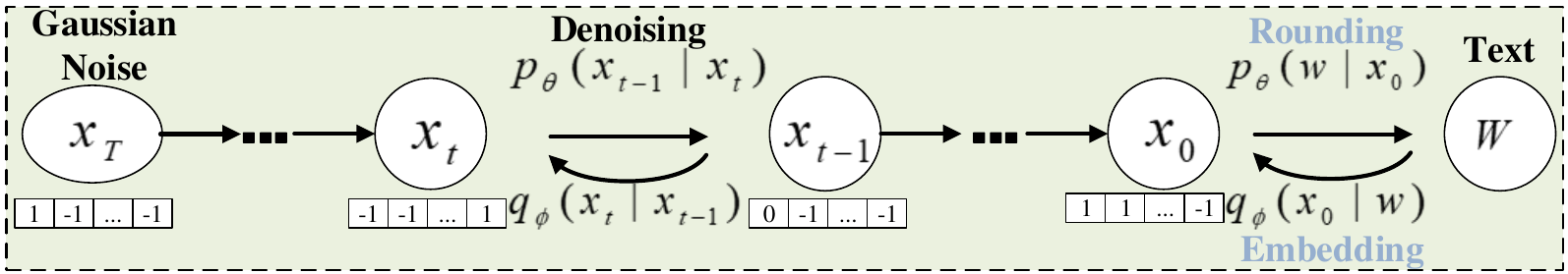}
\end{center}
\caption{A graphical model representing the forward and reverse diffusion processes. Following the existing research in \cite{li2022diffusion},  a Markov transition is introduced between $x_{0}$ and $w$ to achieve end-to-end training and optimize the discrete space based on a quantization method. The discrete space in $x_t$ will be remodeled with a quantized vector $[-1, -1, ..., 1]$.}
\label{Figure3-2}
\end{figure*}

To address (2) and (3), we propose a novel controllable DLM called Quantized Embedding Controllable Diffusion Language Models (QE-CDLM). This model is based on continuous diffusion and incorporates an adaption FT method to reduce the number of tunable weights. As shown in Figure \ref{Figure3-1}, the QE-CDLM begins with a pre-trained language model (LM), which is fed with Gaussian noise vectors and word-corresponding vectors. The embedding spaces for different control tasks are then remodeled and optimized using corresponding quantization methods. This quantization process on embedding vectors leads to a reduction in the perplexity of the generated text. Additionally, the LoRA method, an efficient fine-tuning technique, balances the trade-off between tunable weights and the quality of DLMs.

The contributions of QE-CDLM are as follows:
\begin{enumerate}
\item We validate that quantizing the task-adapting embedding vectors makes controllable DLMs converging faster than the method of independent learned embedding vectors.

\item The quantized embedding vector optimizes the task-specific embedding space in controllable DLMs. Utilizing quantization techniques, e.g., ternarization and binarization of embedding vectors, leads to control texts with improved perplexity.

\item The adaption fine-tuning method achieves a satisfying trade-off between the quality of controllable DLMs and the number of tunable weights. Compared to classical FT methods, this approach shows competitive performance in generating control texts with a reduced number of tunable weights, making it more affordable, and thus increasing the portability.
\end{enumerate}

Our code to reproduce all the experimental results is publicly available at \textcolor{red}{\url{https://github.com/ChengKang520/Q-Controllable-DLM}}.

\section{Related Work}


\subsection{Diffusion Models for Text Generation}

Diffusion generative models were initially proposed in \cite{sohl2015deep} and have shown state-of-the-art sample quality in image and audio domains, readers of interest can be referred to \cite{ho2020denoising,liu2021dexperts,mittal2021symbolic,nichol2021improved}. In the language domain, DiffusionBERT explores training BERT to learn the reverse process of a discrete diffusion process with an absorbing state \cite{he2022diffusionbert}. Recent work on continuous diffusion and controllable text generation \cite{li2022diffusion} has demonstrated that language models can successfully control simple sentence attributes. In general, a diffusion model \cite{ho2020denoising,nichol2021improved} is a latent variable model that represents data $x_{0} \in \mathbb{R}^{d}$ as a Markov chain $x_{T}...x_{0}$, with the respective variable in $\mathbb{R}^d$, and $x_{T}$ following a Gaussian distribution. The diffusion model aims to denoise the sequence of latent variables $x_{T:1}$ to approximate target samples generated from the target data distribution (see Figure \ref{Figure3-2}). The initial state $p_{\theta}$ is approximated as $\mathcal{N}(0,I)$, and the denoising transition $x_{t} \rightarrow x_{t-1}$ is parameterized by the model $p_{\theta}(x_{t}|x_{t-1}) = \mathcal{N}(x_{t-1};\mu_{\theta}(x_{t},t),\sigma_{\theta}(x_{t},t))$. For instance, $\mu_{\theta}$ and $\sigma_{\theta}$ can be determined using a Transformer \cite{li2022diffusion}.

To train the diffusion model, we use a forward process to generate intermediate latent variables $x_{1:T}$. The process starts by incrementally adding Gaussian noise to the initial data point $x_{0}$. As the diffusion progresses through step $T$, the samples $x_T$ become approximately Gaussian. Each transition $x_{t-1} \rightarrow x_{t}$ is parameterized by $q(x_{t}|x_{t-1})=\mathcal{N}(x_{t};\sqrt{1-\beta_{t}}x_{t-1},\beta_{t}I)$, where $\beta_{t}$ is a hyperparameter representing the amount of noise added at diffusion step $t$. The forward process $q$ does not contain any trainable parameters.

The training objective of the diffusion model is to maximize the marginal likelihood of the data, which is formulated as $\mathbb{E}_{x_{0} \sim p_{data}\{\log[p_{\theta}(x_{0})]\}}$. To achieve this, we utilize the variational lower bound of $\log[p_{\theta}(x_{0})]$ \cite{sohl2015deep}. The model is trained to reverse the diffusion process and accurately reconstruct the original data.
\begin{equation} \label{eq8}
\begin{aligned} &&\mathcal{L}_{\mathrm{VLB}}\left(\mathrm{x}_0\right) \ &= \underset{q\left(\mathrm{x}_{1: T} \mid \mathrm{x}_0\right)}{\mathbb{E}}[\log \frac{q\left(\mathrm{x}_T \mid \mathrm{x}_0\right)}{p_\theta\left(\mathrm{x}_T\right)}\\
&& \ &+\sum_{t=2}^T \log \frac{q\left(\mathrm{x}_{t-1} \mid \mathrm{x}_0, \mathrm{x}_t\right)}{p_\theta\left(\mathrm{x}_{t-1} \mid \mathrm{x}_t\right)}-\log p_\theta\left(\mathrm{x}_0 \mid \mathrm{x}_1\right)].
\end{aligned}
\end{equation}

However, achieving this goal can be challenging and may necessitate the use of several optimization techniques to stabilize it \cite{nichol2021improved}. To tackle this issue, \cite{ho2020denoising} proposed a straightforward surrogate objective. They extended and reweighted each KL-divergence term in $L_{VLB}$ to create a mean-squared error loss, which is also referred to as \cite{li2022diffusion}:
\begin{equation} \label{eq9}
\begin{aligned}
\mathcal{L}_{\text {simple }}\left(\mathrm{x}_0\right)=\sum_{t=1}^T \underset{q\left(\mathrm{x}_t \mid \mathrm{x}_0\right)}{\mathbb{E}}\left\|\mu_\theta\left(\mathrm{x}_t, t\right)-\hat{\mu}\left(\mathrm{x}_t, \mathrm{x}_0\right)\right\|^2.
\end{aligned}
\end{equation}

where $\hat{\mu}\left(\mathrm{x}_t, \mathrm{x}_0\right)$ is the mean of the posterior $q(x_{t-1}|x_{0},x_{t})$ which is a closed form Gaussian, and $\mu_{\theta}(x_{t},t)$ is the predicted mean of $p_{\theta}(x_{t-1}|x_{t})$ computed by a neural network. We also make use of similar simplifications in DLM \cite{li2022diffusion} to stabilize training and improve sample quality.

\subsection{Controllable Text Generation}

Controllable text generation (CTG) is the task of generating text while adhering to a given controlled element \cite{prabhumoye2020exploring}. In plug-and-play controllable generation, the LM is kept frozen, and its output is constrained using potential functions, such as classifiers \cite{pascual2021plug, dathathri2019plug}. FUDGE \cite{yang2021fudge} focuses on the partial sequence and reweights the LM prediction at each token based on an estimate of control satisfaction. DExperts \cite{liu2021dexperts} achieves control by reweighting the LM prediction at each token using a smaller fine-tuned or trained LM for the specific control task. Diffusion-LM \cite{li2022diffusion} fine-tunes the pre-trained weights while updating them based on a controllable text generation algorithm. In CTG, given a vocabulary $\mathbi{V}$, the goal is to generate a target text $X = \left\{ x_0, x_1, \ldots, x_t, \ldots, x_T \right\}$, where $x_t \in \mathbi{V}$, while taking into account a control element denoted as $w$. Formally, CTG can be expressed as follows:
\begin{gather} \label{eq1}
P(X \mid w)=p\left(x_0, x_1, \ldots, x_T \mid w\right).
\end{gather}
The specific expression of $w$ may vary in accordance with different tasks. For the sentence $Y$ generated by the model, it is also expected to conform to the constraint conditions and the general natural language characteristics (e.g.,  fluency, rationality, readability, and control success). \cite{yang2021fudge,li2022diffusion}.

To convert a continuous diffusion model into discrete text representation, we employ an embedding function  $\operatorname{EMB}\left(w_i\right)$, which maps each word's embedding space to a corresponding vector in $\mathbb{R}^d$. The embedding vector is represented as $\operatorname{EMB}(\mathbf{w})=\left[\operatorname{EMB}\left(w_1\right), \ldots, \operatorname{EMB}\left(w_n\right)\right] \in \mathbb{R}^{n d}$, where $n$ is the length of the text. In the forward process, the diffusion model introduces a Markov transition from discrete words $\mathrm{w}$ to $\mathrm{x}_0$, which is parameterized by $q_\phi\left(\mathrm{X}_0 \mid \mathbf{w}\right)=\mathcal{N}\left(\operatorname{EMB}(\mathbf{w}), \sigma_0 I\right)$. In the reverse process, a trainable rounding step is added, which is parameterized by $p_\theta\left(\mathbf{w} \mid \mathrm{X}_0\right)=\prod_{i=1}^n p_\theta\left(w_i \mid x_i\right)$. This rounding step optimizes the connection between the continuous $x_{0}$ and discrete text, where $p_\theta\left(w_i \mid x_i\right)$ is modeled as a Softmax distribution. The training objective introduced above is now updated according to:

\begin{equation}
\begin{aligned} \label{eq10}
&&\mathcal{L}_{\mathrm{VLB}}\left(\mathbf{w}\right) \ &= \underset{q\left(\mathrm{x}_{0: T} \mid \mathbf{w}\right)}{\mathbb{E}}[\underbrace{\log \frac{q\left(\mathrm{x}_T \mid \mathrm{x}_0\right)q\left(\mathrm{x}_0 \mid \mathbf{w}\right)}{p_\theta\left(\mathrm{x}_T\right)}}_{\text{Part A}}\\
&& \ &+\underbrace{\sum_{t=2}^T \log \frac{q\left(\mathrm{x}_{t-1} \mid \mathrm{x}_0, \mathrm{x}_t\right)q\left(\mathrm{x}_0 \mid \mathbf{w}, \mathrm{x}_t\right)}{p_\theta\left(\mathrm{x}_{t-1} \mid \mathrm{x}_t\right)}}_{\text{Part B}} \\
&& \ &-\underbrace{\log \left(p_\theta\left(\mathbf{w} \mid \mathrm{x}_0\right)p_\theta\left(\mathrm{x}_0 \mid \mathrm{x}_1\right )\right)}_{\text{Part C}}].
\end{aligned}
\end{equation}

Then we get

\begin{equation}  \label{eq11}
\begin{aligned}
&&\mathcal{L}_{\mathrm{VLB}}\left(\mathbf{w}\right) &=\underset{q_\phi\left(\mathbf{x}_{0: T} \mid \mathbf{w}\right)}{\mathbb{E}}[\underbrace{\log \frac{q\left(\mathbf{x}_T \mid \mathbf{x}_0\right)}{p_\theta\left(\mathbf{x}_T\right)}}_{L_T}\\
&& \ &+\underbrace{\sum_{t=0}^T \log q\left(\mathrm{x}_0 \mid \mathbf{w}, \mathrm{x}_t\right)}_{L_w+L_{\texttt{round}}}\\
&& \ &+\sum_{t=1}^T \underbrace{\log \frac{q\left(\mathbf{x}_{t-1} \mid \mathbf{x}_0, \mathbf{x}_t\right)}{p_\theta\left(\mathbf{x}_{t-1} \mid \mathbf{x}_t\right)}}_{L_{t-1}+L_{0}}].
\end{aligned}
\end{equation}

We employ the same simplification which transforms $\mathcal{L}_{\text {vlb }} \rightarrow \mathcal{L}_{\text {simple }}$ to transform $\mathcal{L}_{\text {vlb }}^{\mathrm{e} 2 \mathrm{e}} \rightarrow \mathcal{L}_{\text {simple }}^{\mathrm{e} 2 \mathrm{e}}$, this result is consistent with \cite{li2022diffusion,ho2020denoising}, that is:

\begin{equation}
\underset{q_\phi\left(\mathbf{x}_{0: T} \mid \mathbf{w}\right)}{\mathbb{E}}\left[L_T\right]=\mathbb{E}\left[\| \hat{\mu}\left(\mathbf{x}_T ; \mathbf{x}_0\right) \|^2\right],
\end{equation}

\begin{equation}
\underset{q_\phi\left(\mathbf{x}_{0: T} \mid \mathbf{w}\right)}{\mathbb{E}}\left[L_{t-1}+L_{0}\right]=\mathbb{E}\left[\left\|\hat{\mu}\left(\mathbf{x}_t, \mathbf{x}_0\right)-\mu_\theta\left(\mathbf{x}_t, t\right)\right\|^2\right],
\end{equation}

\begin{equation}
\underset{q_\phi\left(\mathbf{x}_{0: T} \mid \mathbf{w}\right)}{\mathbb{E}}\left[L_{w}+L_{round}\right]=\mathbb{E}\left[\left\|\operatorname{EMB}(w)\right\|^2\right].
\end{equation}

Accordingly, the training objective introduced above now becomes:

\begin{equation}
\begin{aligned}
&&\mathcal{L}_{\text {simple }}^{\mathrm{e} 2 \mathrm{e}}(\mathbf{w}) & =\underset{q_\phi\left(\mathrm{x}_{0: T} \mid \mathbf{w}\right)}{\mathbb{E}}\left[\mathcal{L}_{\text {simple }}\left(\mathrm{x}_0\right)\right]\\
&& \ &+\underset{q_\phi\left(\mathrm{x}_{0: T} \mid \mathbf{w}\right)}{\mathbb{E}}\left[(T+1)\left\|\operatorname{EMB}(\mathbf{w})\right\|^2\right].
\end{aligned}
\end{equation}

We get the same formula in accordance to \cite{li2022diffusion} and train the Transformer model to directly predict $\mathrm{x}_0$ via $f_\theta\left(\mathrm{x}_t, t\right)$ , and use the tractable Gaussian posterior $q\left(\mathrm{x}_{t-1} \mid \mathrm{x}_0, \mathrm{x}_t\right)$ to compute the mean of $\mathrm{x}_{t-1}$, conditioned on predicted $\mathrm{x}_0$ and observed $\mathrm{x}_t: \frac{\sqrt{\alpha_{t-1}} \beta_t}{1-\bar{\alpha}_t} \mathrm{x}_0+\frac{\sqrt{\alpha_t}\left(1-\bar{\alpha}_{t-1}\right)}{1-\bar{\alpha}_t} \mathrm{x}_t$.

\begin{equation}
\begin{aligned}
&& \ & \left\|\hat{\mu}\left(\mathrm{x}_t, \mathrm{x}_0\right)-\mu_\theta\left(\mathrm{x}_t, t\right)\right\|^2 \\
&& \ &=  ||\left(\frac{\sqrt{\bar{\alpha}_{t-1}} \beta_t}{1-\bar{\alpha}_t} \mathrm{x}_0+\frac{\sqrt{\alpha_t}\left(1-\bar{\alpha}_{t-1}\right)}{1-\bar{\alpha}_t} \mathrm{x}_t\right)-\\
&& \ &\left(\frac{\sqrt{\bar{\alpha}_{t-1}} \beta_t}{1-\bar{\alpha}_t} f_\theta\left(\mathrm{x}_t, t\right)+\frac{\sqrt{\alpha_t}\left(1-\bar{\alpha}_{t-1}\right)}{1-\bar{\alpha}_t} \mathrm{x}_t\right)||^2 \\
&& \ &= ||\frac{\sqrt{\bar{\alpha}_{t-1}} \beta_t}{1-\bar{\alpha}_t}\left(\mathrm{x}_0-f_\theta\left(\mathrm{x}_t, t\right)\right)||^2 \\
&& \ & \propto \left\|\mathrm{x}_0-f_\theta\left(\mathrm{x}_t, t\right)\right\|^2.
\end{aligned}
\end{equation}

\subsection{Vector Quantization Methods}

\begin{figure*}
\begin{center}
\includegraphics[scale=0.32]{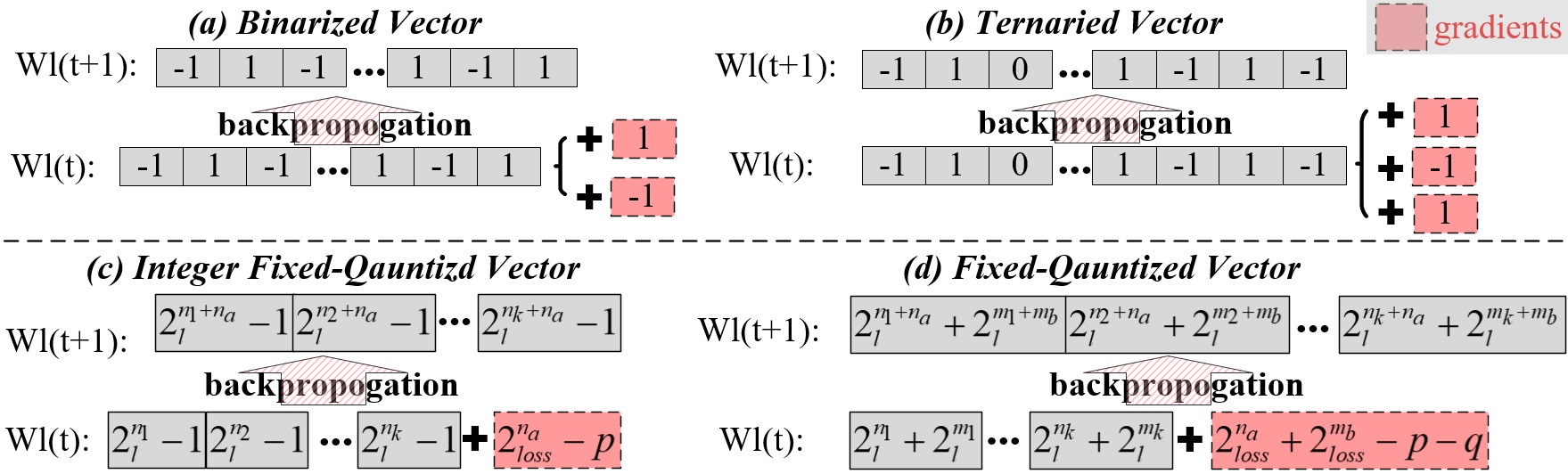}
\end{center}
   \caption{The schematic diagram of four quantization methods in terms of the backpropagation process. For the gradient, $p$ is the remainder in the integer part, and $q$ is the remainder in the fractional part.}
\label{Figure3-3}
\end{figure*}

Quantization methods have been proven to aid in clustering (see \cite{gong2014compressing,tacsdemir2012vector,mavridis2020convergence} and references therein). Previous studies have shown that diffusion models using quantized vectors can effectively accelerate text-to-image generation while maintaining balanced quality \cite{gu2022vector,bond2022unleashing}. This approach has achieved state-of-the-art (SOTA) results on the SLP evaluation benchmark \cite{xie2022vector}, and it offers an appealing trade-off between compression ratio and accuracy \cite{park2022nuqmm}. To deal with the growing model size and the increasing computational and memory requirements, research has explored text diffusion models with quantized vectors, enabling the use of extremely low precision (e.g., 1-bit) weights and activations, recent work can be referred to, e.g., \cite{hubara2017quantized,gu2022vector,croitoru2022diffusion}. In this paper, we study the quantization of the embedding vectors of DLMs on corresponding control tasks so as to enhance the generation process and optimize the task-specific embedding space.

There are four vector quantization methods: (1) Binary, (2) Ternary, (3) Points Quantization, and (4) Fixed-Points Quantization. We applied these four quantization methods to our fine-tuning tasks. To reduce the number of computing bits, we quantized the embedding vector from 8-bit to binary bits. In Figure \ref{Figure3-3}, the update gradients should also be quantized to accelerate the inference process.

\subsubsection{Binarized Vectors} 
The binarized vectors impose constraints on both the weights and activations, limiting them to either $+1$ or $-1$ \cite{hubara2017quantized,courbariaux2015binaryconnect,tu2022adabin}. These binary values offer hardware advantages and are straightforward to implement, yielding effective results in practice:
\begin{equation}
x_b = \text{clip}\left(\frac{x+1}{2}, -1,1\right)=\max(-1, \min(1, \frac{x+1}{2})),
\end{equation}
where $x$ is the real-valued variable.

\subsubsection{Ternarized Vectors} 
Derived from binarized vectors, ternarized vectors \cite{heinrich2018ternarynet} employ an additional bit to represent the value $0$. Neural networks utilizing residual quantization through ternarization of vectors have shown promising results in achieving high recognition accuracy and faster processing times \cite{li2021trq}. Ternarized $(x)$ is defined as follows:
\begin{gather}
\text{tern}(x)= \begin{cases}+1 & \text { if } x>0.5 \\ 0 & \text { if }|x| \leq 0.5 \\ -1 & \text { else }\end{cases}.
\end{gather}

\subsubsection{Quantized Vectors} To quantize the weights of the convolutional layer and the linear layer, We followed the quantization schemes suggested by \cite{hubara2017quantized}:
\begin{equation}
\text{quant}(x) = \text{clip} \left(\mathsf{F_{c}} \cdot \left(\frac{x}{2^{\text{n}}-1}\right) \times 2^{\text{n-1}}, V_{\min}, V_{\max}\right).
\end{equation}

where $V_{\min}$ and $V_{\max}$ are the minimum and maximum scale range respectively. $\mathsf{F_{c}}$ is the quantization function, and it can be a round function. 

\subsubsection{Fixed-Points Quantized Vectors}. The study utilized a fixed-point quantization neural network \cite{askarihemmat2019u} to expedite the segmentation of medical images by implementing a reconstructed U-Net \cite{ronneberger2015u}. The fixed-point quantization function was applied to quantize the parameters (weights and activation) and optimize the process:
\begin{gather} \label{eq5}
\begin{split}
& \text{fixed-quant}(x, n) = (\mathsf{F_{c}}(\text{clip}\left(x, 0, 2^n - 1\right) \ll n)) \gg n
\end{split},
\end{gather}
where $\mathsf{F_{c}}$ is the round function which projects its input to the nearest integer, and $\ll$ and $\gg$ are shift left and right operators, respectively. In our simulation, shift left and right are implemented by multiplication and division in power of $2$. The quantizer here uses bits shift operation which quantizes as an input $x \in \mathbb{R}$ to the closest value that can be represented by $n$ bits.

\subsection{Parameter-Efficient Fine-Tuning} 
Many studies have addressed the issue of redundant knowledge in fully pre-trained language models for downstream tasks in natural language processing (NLP) domains. They have achieved this by adapting specific vectors or learning extra weights while freezing most of the pre-trained weights on new downstream applications. Prior to deploying the model, extra task-specific parameters are loaded for the respective downstream task and associated with the pre-trained weights to achieve better operational efficiency. LoRA \cite{hu2021lora} is a method that attains this and successfully tackles the challenge of inference latency \cite{pfeiffer2020adapterfusion,houlsby2019parameter}. This approach allows for extending the depth of models or reducing the models' usable sequence length \cite{he2021towards,li2021prefix, lester2021power} to balance the trade-off between efficiency and model quality \cite{hu2021lora}. Finally, LoRA is adopted to fine-tune the proposed models using a pre-trained LM, such as BERT \cite{devlin2018bert}.

\section{Problem States}

In this section, we present three main problems of controllable DLMs. 

\subsection{The Complex Embedding Space of Diffusion LMs}

\begin{figure}[h]
\begin{center}
\includegraphics[scale=0.7]{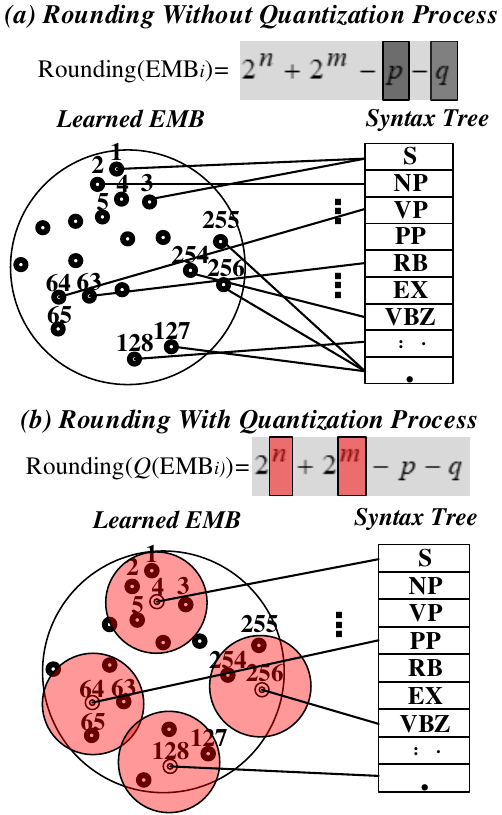}
\end{center}
   \caption{The rounding difference between with and without the quantization processing.}
\label{Figure3-4}
\end{figure}

The learned embedding in this context refers to a mapping from discrete text to the continuous variable $\mathrm{x}_0$. The process involves using $\operatorname{\argmax} p_\theta\left(\mathbf{w} \mid \mathrm{x}_0\right)=\prod_{i=1}^n p_\theta\left(w_i \mid x_i\right)$ to round a predicted $\mathrm{x}_0$ back to discrete text by selecting the most probable word for each respective position \cite{li2022diffusion}. This argmax-rounding effectively converts the embedding space back to discrete text, as the denoising steps ensure that $\mathrm{x}_0$ precisely represents the embedding of some words \cite{li2022diffusion}. It is important to note that generating a single word via $\mathrm{x}_0$ is unlikely, as it cannot be measured based on perplexity and other metrics. Additionally, pre-trained LMs contain task-independent knowledge that does not contain specific task-relevant embedding spaces. As shown in Figure \ref{Figure3-4}, when controller is the \textit{Syntax Tree}, the number of end notes is extremely smaller than that of tokens. Consequently, when the rounding process is used to optimise the embedding space, there is still a greater perplexity if using original embedding space. Obviously, a larger embedding space could lead to high computational cost, and the convergence speed of controllable DLMs could also become slower. While the quantization rounding process can optimize the embedding space, it can compress the task-relevant space to a smaller, better-adapted one. As a result, the main challenge lies in optimizing the embedding space, especially when dealing with a large space.

\subsection{Efficiency and Effectiveness of the Fine-Tuning}

During the full fine-tuning, we initialize the pre-trained parameters $\Phi_{0}$, which are then updated to $\Phi$. More specifically, $P_{\Phi}$ represents the pre-trained weights, and $\Delta \Phi$ represents the trainable language model distribution. Gradient updates are performed on the following log-likelihood objective:
\begin{gather} \label{eq2}
\underset{\Theta}{\max}\sum_{(x,y) \in \mathbb{Z}}^{}\sum_{t=1}^{|y|} \log(P_{\Phi} + \Delta \Phi (\Theta)(y_{t}|x,y_{<t})).
\end{gather}

During the inference process, the pre-trained weights are frozen, and the adaptation matrices are incorporated into the frozen model to approximate the update of the pre-trained weights. However, this introduces an additional computing load on the inference process. Alternatively, the extra tunable parameters can be computed in parallel using a parallel architecture, which can help maintain the speed of the inference process even with frozen pre-trained weights and fine-tuning of the extra tunable parameters. Hence, the challenge is to design a parallel tunable module that preserves or even enhances the fine-tuning performance.

\subsection{Theoretical Inference Speed}

The inference step refers to the process of generating text based on the architecture of the model. Traditionally, fine-tuning pre-trained LMs has been used to adapt the embedding space for specific tasks, but this approach retains the original architecture of the model. Recent research has explored the use of quantization vectors to speed up the inference process, as they involve lighter computations compared to full-fledged LMs \cite{yu2021vector,zafrir2019q8bert,yao2023comprehensive}. However, a challenge arises when adopting quantized vectors, as it requires pre-training the model from scratch on large language datasets to acquire the necessary knowledge. This raises the question of whether there are alternative approaches that can retain the benefits of pre-trained weights while quantizing certain parts of the LMs.

\section{Quantized Embedding Vectors for Controllable Diffusion LMs}

\subsection{Controllable Generation and Decoding of Diffusion-LM} 

\subsubsection{Controllable Text Generation} Based on the approach presented in \cite{li2022diffusion} for controlling text generation, we also adopt a method that controls the sequence of continuous latent variables $x_{0:T}$ defined by Diffusion-LM. Controlling $x_{0:T}$ is equivalent to decoding from the posterior $p\left(\mathrm{X}_{0: T} \mid \mathbf{c}\right)=\prod_{t=1}^T p\left(\mathrm{X}_{t-1} \mid \mathrm{X}_t, \mathbf{c}\right)$. To simplify $p\left(\mathrm{X}_{t-1} \mid \mathrm{X}_t, \mathbf{c}\right) \propto p\left(\mathrm{X}_{t-1} \mid \mathrm{X}_t\right) \cdot p\left(\mathbf{c} \mid \mathrm{X}_{t-1}, \mathrm{X}_t\right)$, we make use of conditional independence assumptions based on prior work on controlling diffusions \cite{song2020score}. Finally, we run a gradient update on $x_{t-1}$ in the $t$-th step:

\begin{equation}\label{eq10c}
\begin{aligned}
 \nabla_{\mathrm{x}_{t-1}} \log p\left(\mathrm{x}_{t-1} \mid \mathrm{x}_t, \mathbf{c}\right)&= \nabla_{\mathrm{x}_{t-1}} \log p\left(\mathrm{x}_{t-1} \mid \mathrm{x}_t\right)\\&+\nabla_{\mathrm{x}_{t-1}} \log p\left(\mathbf{c} \mid \mathrm{x}_{t-1}\right),
\end{aligned}
\end{equation}
where $\log p\left(\mathrm{x}_{t-1} \mid \mathrm{x}_t\right)$ and $\log p\left(\mathbf{c} \mid \mathrm{x}_{t-1}\right)$ is respectively parametrized by the Diffusion-LM and a classifier.

\emph{Fluency regularization} is incorporated in the form of $\lambda s\log p\left(\mathrm{X}_{t-1} \mid \mathrm{X}_t\right) + \log p\left(\mathbf{c} \mid \mathrm{X}_{t-1}\right)$, where $\lambda$ is a hyperparameter that controls the tradeoff between fluency (i.e., the first term) and control (i.e., the second term), based on which we can generate text that is both fluent and controllable. Unlike existing diffusions methods for controllable generation, our approach includes the $\lambda s\log p\left(\mathrm{X}_{t-1} \mid \mathrm{X}_t\right)$ term in the objective, which has been verified to be beneficial for generating text with proper grammar and fluency. Our proposed controllable generation procedure can be treated as a stochastic decoding method that balances the maximization and sampling of $\log p\left(\mathbf{c} \mid \mathrm{X}_{t-1}\right)$, similar to popular text generation techniques like nucleus sampling \cite{holtzman2019curious} or sampling with low temperatures. To manage the computational cost, we apply $3$ steps of the Adagrad update for each diffusion step and downsample the diffusion steps from $2000$ to $200$ following the approach used in Diffusion-LM \cite{li2022diffusion}.

\subsubsection{Minimum Bayes Risk Decoding} The task of conditional text generation, such as machine translation, sentence infilling, and answering, involves generating a single high-quality sequence as the output. To achieve this, we employ Minimum Bayes Risk (MBR) decoding \cite{kumar2004minimum}, which involves aggregating a set of samples $\mathcal{S}$ generated from Diffusion-LM. The goal is to select the sample that minimizes the expected risk according to a specific loss function $\mathcal{L}$ (e.g., negative BLEU score). Mathematically, this can be expressed as $\operatorname{\argmin}{\mathbf{w} \in S} \sum{\mathbf{w}^{\prime} \in S} \frac{1}{|S|} \mathcal{L}\left(\mathbf{w}, \mathbf{w}^{\prime}\right)$. MBR decoding tends to yield high-quality outputs as low-quality samples will differ significantly from the remaining samples and receive penalties from the loss function \cite{li2022diffusion}.

\subsection{Optimize the Embedding Space of Controllable Diffusion LMs}

 $\mathcal{L}_{\text {simple }}^{\mathrm{e} 2 \mathrm{e}}(\mathbf{w})$ is derived from $\mathcal{L}_{\mathrm{vlb}}^{\mathrm{e} 2 \mathrm{e}}(\mathbf{w})$ following the simplification in Appendix A, as well as derivation details. Since the embedding function should be trained to satisfy the control tasks, $q_{\phi}$ contains trainable weights while using the parameterization trick \cite{kingma2013auto,rezende2014stochastic} to back-propagate through sampling steps. In Figure \ref{Figure3-3}, the learned word-embeddings cluster with the same part-of-speech tags (syntactic role) tend to be clustered.

In the respective term of the objective (e.g., POS), it is effective to perform re-parameterization on $\mathcal{L}_{\text{simple }}$ \cite{li2022diffusion} to force Diffusion-LM to explicitly model $\mathrm{x}_0$ . By parametrizing $\mathrm{x}_0$, an analogue to $\mathcal{L}_{\text {simple }}$ can be derived from $\mathcal{L}_{\mathrm{x}_0 \text {-simple }}^{\text {e2e }}\left(\mathrm{x}_0\right)=\sum_{t=1}^T \mathbb{E}_{\mathrm{x}_t}\left\|f_\theta\left(\mathrm{x}_t, t\right)-\mathrm{x}_0\right\|^2$, where the proposed model $f_\theta\left(\mathrm{x}_t, t\right)$ predicts $\mathrm{x}_0$ directly. Predicting $\mathrm{x}_0$ and $\mathrm{x}_{t-1}$ is equivalent to scaling constants as the distribution of $\mathrm{x}_{t-1}$ can be obtained in a closed form through the forward process $\mathrm{x}_{t-1}=\sqrt{\bar{\alpha}} \mathrm{x}_0+\sqrt{1-\bar{\alpha}} \epsilon$, see Appendix A for more detail. The aforementioned process forces the neural network to predict $\mathrm{x}_0$ in the respective term. Lastly, models trained with this objective are capable of precisely inducing $\mathrm{x}_0$ centered at a word embedding.

In the standard generation method, the model denoises $\mathrm{x}_t$ to $\mathrm{x}_{t-1}$ by first computing an estimate of $\mathrm{x}_0$ via $f_\theta\left(\mathrm{x}_t, t\right)$ and then sampling $\mathrm{x}_{t-1}$ conditioned on this estimate: $\mathrm{x}_{t-1}=\sqrt{\bar{\alpha}} f_\theta\left(\mathrm{x}_t, t\right)+\sqrt{1-\bar{\alpha}} \epsilon$, where $\bar{\alpha}_t=\prod_{s=0}^t\left(1-\beta_s\right)$ and $\epsilon \sim \mathcal{N}(0, I)$. In the clamping trick, the model maps the predicted vector $f_\theta\left(\mathrm{x}_t, t\right)$ to its nearest word embedding sequence. The sampling step is converted as $\mathrm{x}_{t-1}=\sqrt{\bar{\alpha}} \cdot \text{clip}\left(f_\theta\left(\mathrm{x}_t, t\right), 0, 2^n - 1\right)+\sqrt{1-\bar{\alpha}} \epsilon$. The clamping trick propels $f_\theta\left(\mathrm{x}_t, t\right)$ to a proper embedding space and forces the predicted vector to commit to a target word during diffusion intermediate steps \cite{li2022diffusion}. To improve precision and reduce rounding errors, this study suggests using quantization methods during the clamping process. Quantized embedding vectors are closer to the rounding space that represents the distribution of the most probable word position (or term position). In previous research \cite{li2022diffusion}, the clamping trick was applied for all diffusion steps and could be set as a hyperparameter at the starting step of the clamping process. The rounding method maps quantized vectors in the embedding space back to words. After jointly learning the diffusion model parameters and word embeddings, the results indicate that the use of quantized vectors simplifies the embedding space and optimizes perplexity.

\subsection{Quantize Embedding Vectors}

Previous DLMs replaced the pre-trained embedding vector with a task-specific embedding vector and used a rounding process to generate discrete text \cite{li2022diffusion}. However, a general embedding vector with 64 or 16 bits may lead to unsatisfactory convergence results, as the larger discrete space created by a 64-bit pr 16-bit processor may not be ideal compared to an embedding vector with only eight bits. To address this issue, four quantization methods (Figure \ref{Figure3-6} in Appendix B) are proposed to quantize the embedding vector, focusing on processing each bit separately. The quantization process is defined as follows:

\begin{gather} 
Q_n= 2^{s \times \mathsf{Round} (s \times \log_{2}x )}.
\end{gather}

Here, $x$ represents the input, and $s \in {-1, 1}$ is chosen to select the integer or fractional part. $n \in \mathbb{N}$ represents the number of bits, and when $s=-1$, the fractional part of the embedding vector is quantized. The quantization method on DLMs is formulated as:
\begin{equation} \label{eq3}
\tilde{x}_n = \mathsf{\max}(V_{\min}, \mathsf{\min}(V_{\max}, \mathbf{Q_n})).
\end{equation}

Where $V_{\min}$ and $V_{\max}$ denote the minimum and maximum scale range, respectively. $\mathbf{Q_n}$ represents the quantization result computed by the corresponding quantization method. In Table \ref{Table3-3} and Table \ref{Table3-4}, Q$0$i$\beta$f indicates that the fractional part with $\beta$ bits is quantized, and Q$\alpha$i$0$f suggests that the integer part with $\alpha$ bits is quantized. Finally, the model maps the predicted vector $f_\theta\left(\mathrm{x}_t, t\right)$ to a quantized nearest word embedding sequence:

\begin{gather} \label{eq4}
\mathrm{x}_{t-1}=\sqrt{\bar{\alpha}} \cdot \text{clip}\left(\text{quant}(f_{\theta}\left(\mathrm{x}_t, t\right)\right), 0, 2^n - 1)+\sqrt{1-\bar{\alpha}} \epsilon.
\end{gather}

\begin{figure*}
\begin{center}
\includegraphics[scale=0.3]{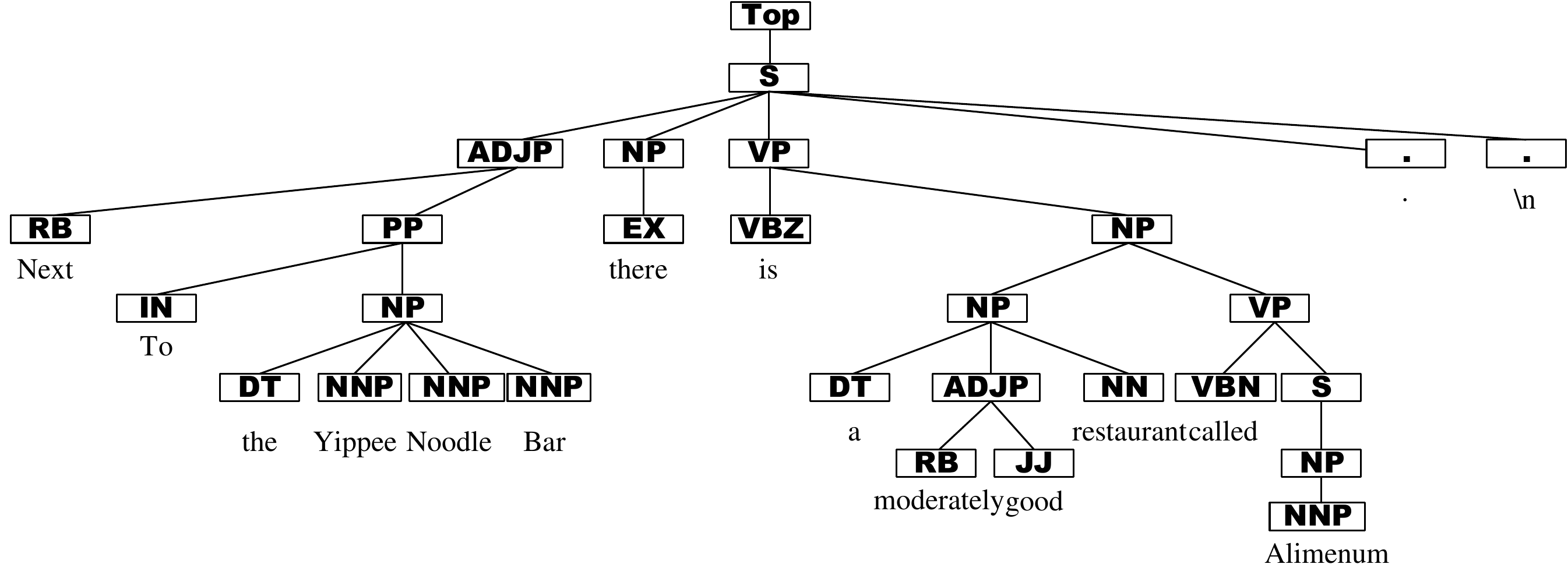}
\end{center}
   \caption{The structure of the Syntax Tree presented in Table \ref{Table3-2}.}
\label{Figure3-5}
\end{figure*}

\subsection{Adaption Fine-Tune on Diffusion LMs} 

In this part, LoRA \cite{hu2021lora} is applied to any subset of weight matrices to reduce the number of trainable parameters. There are four weight matrices in the self-attention module ($W_q$, $W_k$, $W_v$, and $W_o$) and two in the multilayer perceptron (MLP) module. For simplicity and parameter-efficiency, $W_q \in \mathbb{R}^{M \times d}$ (or $W_k$, $W_v$)  is re-scaled with these two trainable matrices $W_{down} \in {R^{d\times{r}}}$ and $W_{up} \in {R^{r\times{k}}}$, where $M$ denotes the length of the input sequence, and $D$ is the dimension of hidden states. The model updates the $key$ and $value$ projection matrices ($W_{k}, W_{v}$) in the multi-head attention sub-layer by multiplying with the above-described two tunable matrices $W_{down}$ and $W_{up}$. In terms of a specific hidden input $H$, LoRA modifies the projection output $H_{o}$ as:
\begin{equation}
    H_{o}\leftarrow H_{o}+s \cdot f(HW_{down})W_{up}.
\end{equation}
where $s \geqslant 1$ denotes a tunable scalar hyperparameter. $f$ expresses the activaiton funciton.

\section{Experiments}

Diffusion-LM is trained on three language modelling tasks under three quantization methods, and the controllable generation method is adopted to generate the text under five classifier-guided controllers. Lastly, the LoRA FT method is employed to fine-tune the pre-trained LM.

\subsection{Datasets and Experiment Settings} 

This study employs three datasets for fine-tuning Diffusion-LM: E2E \cite{novikova2017e2e}, ROCStories \cite{mostafazadeh2016corpus}, and WikiText \cite{merity2016pointer}. The E2E dataset consists of 50K restaurant reviews labeled by eight fields, such as food type, price, and customer rating. The ROCStories dataset includes 98K five-sentence stories, capturing various causal and temporal commonsense relations between daily events. The WikiText language modeling dataset contains over 100 million tokens extracted from verified Good and Featured articles on Wikipedia, comprising WikiText2 and WikiText103.

In this study, the Diffusion-LM with quantized embedding vectors follows the BERT \cite{devlin2018bert} architecture, with $80$M parameters, a sequence length of $n = 64$, and $2000$ diffusion steps. For fair comparison, the embedding dimension is set to $d = 16$ for E2E, $d = 128$ for ROCStories and WikiText2, and $d = 1024$ for WikiText103. Refer to Appendix F for hyperparameter details. During decoding, we downsample to $200$ diffusion steps for E2E, while maintaining $2000$ steps for ROCStories and WikiText. Additionally, the input-up, time-embedding, and output-down vectors are binarized, ternarized, and fixed-quantized to evaluate the performance of different quantization methods. For the LoRA fine-tuning method, the hyperparameters \textit{Adaption} $r_q=r_v=8$ and \textit{LoRA} $\alpha=16$ are set.

\subsection{Hyperparameters}

\subsubsection{Diffusion-LM hyperparameters} 
For the Quantized Embedding Diffusion-LM, we consider several hyperparameter settings. The number of diffusion steps is set to 2000, and we use the BERT-base architecture \cite{devlin2018bert} with a sequence length of 64. The embedding dimensions $d$ are chosen based on the dataset: $d=16$ for the E2E dataset, $d=128$ for ROCStories and WikiText2, and $d=1024$ for WikiText103. Regarding the noise schedule, we use the sqrt schedule (as described in Appendix A.1). It's worth noting that the advantage of the sqrt schedule is not prominent when we use the $x0$-parametrization \cite{li2022diffusion}. For the quantization process, denoted as $\text{quant} = {\mathbf{bnn}, \mathbf{tern}, \textbf{Q}\alpha \textbf{i} \beta \textbf{f}}$, we experiment with different values of $\alpha$ and $\beta$. Specifically, we select $\alpha$ from ${0, 4, 8, 16}$ and $\beta$ from ${0, 4, 8, 16}$ for the E2E, ROCStories, and WikiText datasets. These quantization processes have been shown to be faster and more robust in differentiating parametrizations and embedding dimensions, as depicted in Figure \ref{Figure3-6}. However, we avoid using $\alpha \neq \beta = 0$, as $\beta=0$ implies that no quantization method is applied to the corresponding part.

\subsubsection{Training hyperparameters} 
We train Diffusion-LMs using AdamW optimizer and a linear decay learning rate starting at 1e-4, dropout of 0.1, batch size of 64, and the total number of training iteration is 200K for the E2E dataset, 800K for the ROCStories and WikiText2 datasets and 1.2M for the WikiText103 dataset. Our Diffusion-LMs are trained on a single GPU, and it takes approximately 3 hours to train for 200K iterations on a single Tesla A100 GPU with 40GB graphic memory.


\subsubsection{Controllable Generation hyperparameters} 
To achieve controllable generation, we run gradient update on the continuous latents of Diffusion-LM. We use the AdaGrad optimizer \cite{duchi2011adaptive} to update the latent variables, and we tune the learning rate, $\operatorname{lr} \in\{0.05,0.1,0.15,0.2\}$, the trade-off parameter $\lambda \in\{0.1,0.01,0.001,0.0005\}$ (found in Appendix A.3), the number of quantized bits $q_{n} \in \{0, \alpha, \beta\}$, and the part should be quantized $c \in \{-1, 1\}$.

\begin{table}[t]
\caption{Hyperparameters for controllable generation methods. }
\label{Table3-7}
\centering
\scriptsize
\setlength\tabcolsep{0.8pt}
\begin{tabular}{c|cccc|cccc|cccc|cccc|cccc}
\toprule
\multirow{2}{*}{EMB}                & \multicolumn{4}{c|}{Semantic}           & \multicolumn{4}{c|}{Part-of-speech}        & \multicolumn{4}{c|}{Syntax Tree}          & \multicolumn{4}{c|}{Syntax Spans}       & \multicolumn{4}{c}{Lentgh}              \\
& $\lambda$  & $lr$  & $q_n$                  & $c$  & $\lambda$    & $lr$   & $q_n$                  & $c$   & $\lambda$    & $lr$  & $q_{n}$                  & $c$   & $\lambda$ & $lr$   & $q_{n}$                  & $c$   & $\lambda$  & $lr$  & $q_{n}$                  & $c$  \\ \hline
original                     & 0.01 & 0.1 & -                     & -  & 0.0005 & 0.05 & -                     & -  & 0.0005 & 0.2 & -                     & -  & 0.1 & 0.15 & -                     & -  & 0.01 & 0.1 & -                     & -  \\
Q$\alpha$i$0$f & 0.01 & 0.1 & $\alpha$ & 1  & 0.0005 & 0.05 & $\alpha$ & 1  & 0.0005 & 0.2 & $\alpha$ & 1  & 0.1 & 0.15 & $\alpha$ & 1  & 0.01 & 0.1 & $\alpha$ & 1  \\
Q$0$i$\beta$f  & 0.01 & 0.1 & $\beta$  & -1 & 0.0005 & 0.05 & $\beta$  & -1 & 0.0005 & 0.2 & $\beta$  & -1 & 0.1 & 0.15 & $\beta$  & -1 & 0.01 & 0.1 & $\beta$  & -1 \\
ternary                      & 0.01 & 0.1 & 2                     & 1  & 0.0005 & 0.05 & 2                     & 1  & 0.0005 & 0.2 & 2                     & 1  & 0.1 & 0.15 & 2                     & 1  & 0.01 & 0.1 & 2                     & 1  \\
binary                       & 0.01 & 0.1 & 1                     & -1 & 0.0005 & 0.05 & 1                     & -1 & 0.0005 & 0.2 & 1                     & -1 & 0.1 & 0.15 & 1                     & -1 & 0.01 & 0.1 & 1                     & -1 \\ \bottomrule
\end{tabular}
\end{table}

\subsection{Control Tasks}

\begin{figure*}
\centering
\subfigure[Fully FT with quantized embedding vectors (from left to right): no-quatization, fix-quantized fractional 8-bit, fix-quantized fractional 4-bit, binnary.]{
    \centering
    \begin{minipage}[b]{1\textwidth}
        \centering
        \includegraphics[scale=0.35]{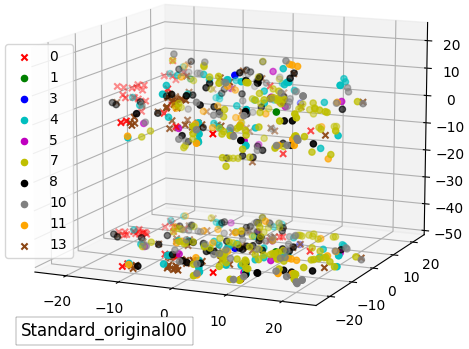}
        \includegraphics[scale=0.35]{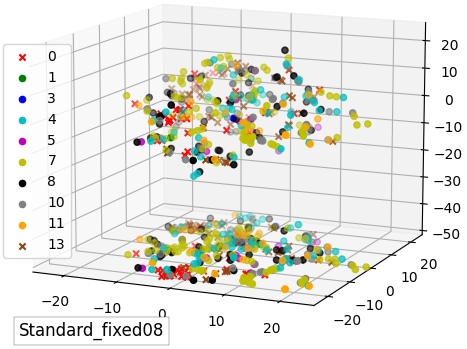} 
        \includegraphics[scale=0.35]{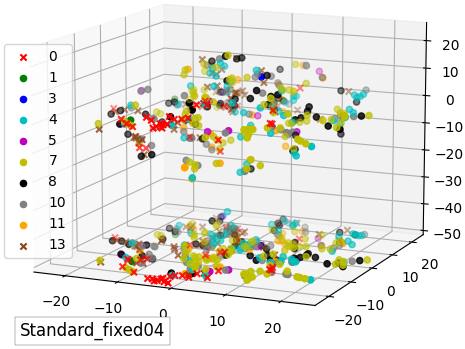}
        \includegraphics[scale=0.35]{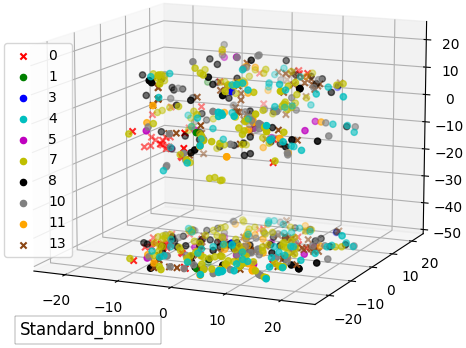}
    \end{minipage}
}
\subfigure[LoRA FT  with quantized embedding vectors (from left to right): no-quatization, fix-quantized fractional 8-bit, fix-quantized fractional 4-bit, binnary.]{
    \begin{minipage}[b]{1\textwidth}
        \centering
        \includegraphics[scale=0.35]{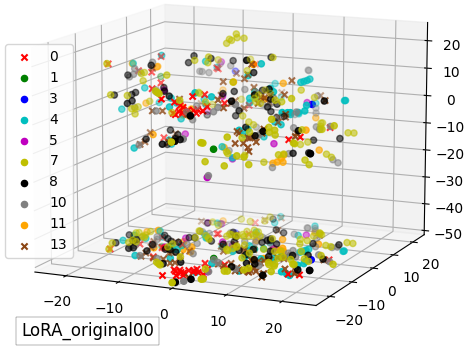} 
        \includegraphics[scale=0.35]{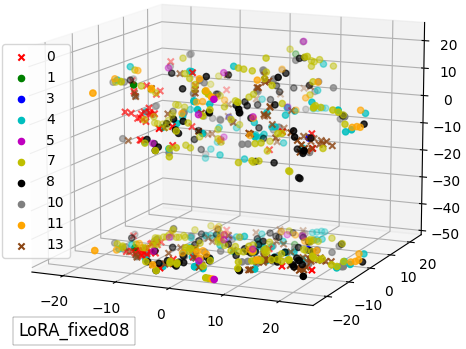} 
        \includegraphics[scale=0.35]{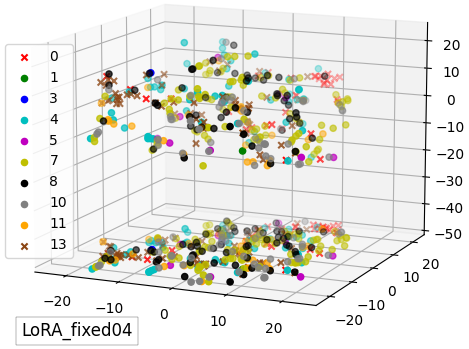}
        \includegraphics[scale=0.35]{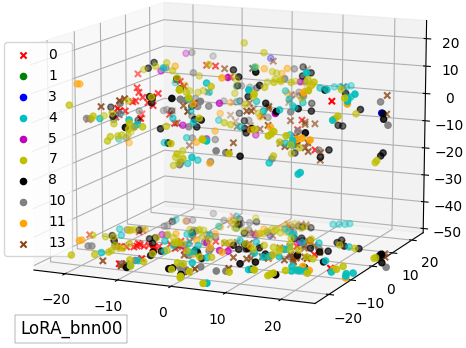}
    \end{minipage}
}
\caption{A 3D t-SNE \cite{van2008visualizing} plot (with its 2d projected plot) of the learned word embeddings. The tag list is [0: 'START', 1: 'ADJ', 2: 'ADV', 3: 'INTJ', 4: 'NOUN', 5: 'PROPN', 6: 'VERB', 7: 'ADP', 8: 'AUX', 9: 'CCONJ', 10: 'DET', 11: 'NUM', 12: 'PART', 13: 'PRON', 14: 'SCONJ', 15: 'PUNCT', 16: 'SYM', 17: 'X', 18: 'END']}
\label{Figure3-6}
\end{figure*}

In the experiments conducted by \cite{li2022diffusion}, extensive tests over five control tasks: Semantic Content, Parts-of-speech, Syntax Tree, Syntax Spans, and Length. Each of the first four tasks had a corresponding classifier, while the last task was classifier-free. To perform the evaluation, 200 control targets (e.g., rating=5 star) were sampled from the validation splits, and 50 samples were generated for each control target. To assess the fluency of the generated text, we refer to the existing research by \cite{yang2021fudge,li2022diffusion, dathathri2019plug}; the generated text was fed to a teacher LM, which was a carefully fine-tuned GPT-2 model. The perplexity of the generated text under the teacher LM was reported as a measure of fluency. It's worth noting that prior works \cite{yang2021fudge, dathathri2019plug} used GPT \cite{radford2018improving} as the teacher LM. However, for a fair comparison, a fine-tuned GPT-2 model was employed in this study, as the Diffusion-LM also generates \texttt{UNK} tokens (\texttt{UNK} does not exist in pre-trained vocabularies of GPT \cite{li2022diffusion}). The evaluation used two metrics: \texttt{ctrl}, which denotes the success rate and calculates the exact match rate of the 'value', and \texttt{lm-score}, which expresses the perplexity evaluated by the teacher LM.

\subsection{Baselines}
The proposed method is compared with FUDGE and Diffusion-LM for the five control tasks mentioned above.

\textbf{FUDGE} (Future Discriminator Guided Generation): FUDGE utilizes a future discriminator that takes a prefix sequence as input and predicts whether the complete sequence would satisfy the given constraint or control. During decoding, FUDGE reweighs the language model's predictions based on the discriminator scores \cite{yang2021fudge}.

\textbf{Diffusion-LM}: Diffusion-LM \cite{li2022diffusion} employs a gradient-based approach to guide the text generation process towards outputs that satisfy specific structural and semantic controls. It iteratively updates the continuous latent variables of Diffusion-LM to achieve a balance between fluency and control satisfaction.

\section{Main Results}

We train Diffusion-LMs on the E2E, ROCStories, and WikiText datasets. In Table \ref{Table3-3}, the controllable generation of DLM optimized by quantized embedding vectors achieves better outputs than typical DLMs and other autoregressive LMs. Empirically, regardless of whether LoRA is employed or not, when using the same quantization method, we find that learned embeddings, even with random initialization, cluster in the same distribution trend (e.g., part-of-speech), as shown in Figure \ref{Figure3-3}. For negative log-likelihood (NLL, lower is better), please refer to Table \ref{Table3-3} (E2E) and Table \ref{Table3-4} (ROCStories) in Appendix for the comparison result of Loss and mean-square-error (MSE). The likelihood, MSE results, and language fluency can be improved by quantizing the fractional part of the embedding vector, ternarizing, or binarizing the embedding vector. Without control, DLMs developed with the pre-trained BERT-base and quantized embedding vectors exhibit exceptionally high perplexity (\texttt{lm-score}). However, the results achieved using the controller suggest that the metrics (\texttt{ctrl} and \texttt{lm-score}) can be improved, especially when quantizing the fractional part of embedding vectors. Although the LoRA FT method may not reduce perplexity, the tunable weights are reduced by over 90\%. As shown in Table \ref{Table3-1} in Appendix C, the proposed embedding-quantized Diffusion-LM readily achieves higher \texttt{ctrl} and lower \texttt{lm-score} across all classifier-guided control tasks, outperforming FUDGE \cite{yang2021fudge} and DLM \cite{li2022diffusion} baselines in all five tasks significantly.

\subsection{Convergence}

Fine-tuning pre-trained language models involves two main steps: (1) leveraging knowledge from the pre-trained weights, and (2) adapting the embedding space for specific tasks. However, the large embedding space from the pre-trained language model can hinder the convergence process during fine-tuning, as it requires more time to adapt to a specific downstream task. Therefore, optimizing the embedding space by reducing the dimensionality of the pre-trained embeddings becomes crucial for accelerating convergence.

\begin{table*}[hbt!]
\parbox{.45\linewidth}{
\caption{The loss and mean-square-error (MSE) of Diffusion-LMs using different vector-quantization and fine-tuning methods on E2E dataset. A lower MSE means a faster convergence speed.}
\footnotesize\setlength\tabcolsep{3.5pt}
\centering
\begin{tabular}{p{0.14\linewidth} p{0.25\linewidth} p{0.1\linewidth} p{0.1\linewidth} p{0.1\linewidth} p{0.1\linewidth}}
\toprule
\multirow{2}{*}{\begin{tabular}[c]{@{}c@{}}\textbf{Method} \\ (Vector)\end{tabular}}                                      & \multirow{2}{*}{\begin{tabular}[c]{@{}c@{}}\textbf{Basic}+\textit{}\\ \textit{Extra}\textbf{(M)}\end{tabular}} & \multicolumn{2}{c}{\textbf{Training}} & \multicolumn{2}{c}{\textbf{Evaluating}} \\ \cline{3-6} 
&               & \textbf{Loss}        & \textbf{MSE}       & \textbf{Loss}         & \textbf{MSE}        \\ \hline  
\multicolumn{6}{c}{\textit{\textbf{Fully Fine-Tuning} (Lower is better)}} \\ \hline 
Original & \textbf{86.0}+\textit{4.82}                                                                              & 0.612                & 0.532              & 0.608                 & 0.528               \\ \hline
Q$\alpha$i.$0$f   & \begin{tabular}[c]{@{}c@{}}\textbf{86.0}+\textit{4.82}$*\alpha$ \textit{/ 64}\end{tabular}                                                                             & 0.611                & 0.531              & 0.609                 & 0.530               \\ \hline
Q$0$i.$\beta$f   & \begin{tabular}[c]{@{}c@{}}\textbf{86.0}+\textit{4.82}$*\beta$ \textit{/ 64}\end{tabular}                                                                            & 0.254                & 0.244              & 0.254                 & 0.244               \\ \hline
Ternary  & \textbf{86.0}+\textit{0.151}                                                                               & 0.252                     & 0.242                   & 0.254                      & 0.244                    \\ \hline
Binary  & \textbf{86.0}+\textit{0.075}                                                                               & 0.264                & 0.254              & 0.259                 & 0.247               \\  \hline 

\multicolumn{6}{c}{\textit{\textbf{LoRA Fine-Tuning} (Lower is better)}} \\ \hline
Original & \textit{4.82+0.3}                                                                              & 0.250                & 0.240              & 0.248                 & 0.240               \\ \hline
Q$\alpha$i.$0$f   & \begin{tabular}[c]{@{}c@{}}\textit{4.82}$*\alpha$ / \textit{64+0.3}\end{tabular}                                                                              & 0.265                & 0.264              & 0.255                 & 0.251               \\ \hline
Q$0$i.$\beta$f   & \begin{tabular}[c]{@{}c@{}}\textit{4.82}$*\beta$ / \textit{64+0.3}\end{tabular}                                                                             & 0.268                & 0.257              & 0.271                 & 0.261               \\ \hline
Ternary  & \textit{0.151+0.3}                                                                               & 0.266                     & 0.255                   & 0.269                      & 0.259                    \\ \hline
Binary  & \textit{0.075+0.3}                                                                               & 0.293                & 0.279              & 0.292                 & 0.279               \\
\bottomrule
\end{tabular}
\label{Table3-3} 
}
\hfill
\parbox{.45\linewidth}{
\caption{ The loss and mean-square-error (MSE) of Diffusion-LMs using different vector-quantization and fine-tuning methods on ROCStories dataset. A lower MSE means a faster convergence speed.}
\label{Table2}
\footnotesize\setlength\tabcolsep{3.5pt}
\centering
\begin{tabular}{p{0.14\linewidth} p{0.25\linewidth} p{0.1\linewidth} p{0.1\linewidth} p{0.1\linewidth} p{0.1\linewidth}}
\toprule
\multirow{2}{*}{\begin{tabular}[c]{@{}c@{}}\textbf{Method} \\ (Vector)\end{tabular}}                                      & \multirow{2}{*}{\begin{tabular}[c]{@{}c@{}}\textbf{Basic}+\textit{}\\ \textit{Extra}\textbf{(M)}\end{tabular}} & \multicolumn{2}{c}{\textbf{Training}} & \multicolumn{2}{c}{\textbf{Evaluating}} \\ \cline{3-6} 
&                                                                                     & \textbf{Loss}        & \textbf{MSE}       & \textbf{Loss}         & \textbf{MSE}        \\ \hline  
\multicolumn{6}{c}{\textit{\textbf{Fully Fine-Tuning} (Lower is better)}} \\ \hline                                                                     
Original  & \textbf{86.0}+\textit{4.82}                                                                                    & 0.102                & 0.095              & 0.104                 & 0.096               \\ \hline

Q$\alpha$i.0f  & \begin{tabular}[c]{@{}c@{}}\textbf{86.0}+\textit{4.82}$*\alpha$ / \textit{64}\end{tabular}                                                                                & 0.101                & 0.095              & 0.103                 & 0.096               \\ \hline

Q0i.$\beta$f  & \begin{tabular}[c]{@{}c@{}}\textbf{86.0}+\textit{4.82}$*\beta$ / \textit{64}\end{tabular}                                                                                & 0.058                & 0.056              & 0.057                 & 0.055               \\ \hline

Ternary  & \textbf{86.0}+\textit{0.151}                                                                          & 0.063                     & 0.061                   & 0.052                      & 0.050                    \\ \hline
Binary   & \textbf{86.0}+\textit{0.075}                                                                          & 0.059                & 0.057              & 0.058                 & 0.056               \\ \hline  

\multicolumn{6}{c}{\textit{\textbf{LoRA Fine-Tuning} (Lower is better)}} \\ \hline  
             
Original  & \textit{4.82+0.3}                                                                                   & 0.059                & 0.056              & 0.055                 & 0.054               \\ \hline

Q$\alpha$i.0f  & \begin{tabular}[c]{@{}c@{}}\textit{4.82}$*\alpha$ / \textit{64+0.3}\end{tabular}                                                                                  & 0.059                & 0.057              & 0.054                 & 0.052               \\ \hline
Q0i.$\beta$f  & \begin{tabular}[c]{@{}c@{}}\textit{4.82}$*\beta$ / \textit{64+0.3}\end{tabular}                                                                                  & 0.059                & 0.057              & 0.054                 & 0.052               \\ \hline

Ternary  & \textit{0.151+0.3}                                                                          & 0.059                     & 0.057                   & 0.054                      & 0.052                    \\ \hline
Binary   & \textit{0.075+0.3}                                                                          & 0.060                & 0.058              & 0.055                 & 0.053               \\ 
\bottomrule
\end{tabular}
\label{Table3-4} 
}
\end{table*}

In Table \ref{Table3-3} and Table \ref{Table3-4}, we observe that metrics such as Loss and MSE. The MSE measures the match between the label and the generated output. perform better when quantizing the fractional part of embedding vectors. Fine-tuning using the LoRA method improves the performance (lower Loss and MSE) of both E2E and ROCStories datasets when using original and Q$\alpha$i.$0$f embedding vectors. However, the binarizing and ternarizing processes do not benefit the convergence. During LoRA fine-tuning, we freeze the pre-trained weights and fine-tune only a few additional parameters that are inserted into the embedding layer. These extra parameters assist in approximating the update of the pre-trained embedding, but this method adversely affects the ternary and binary quantization processes.

\subsection{Efficiency: Trainable Parameters and Speed}

We represent $W_q$ (or $W_k$, $W_v$) as a unified matrix with dimensions $d \times d$. In the LoRA fine-tuning method, $r$ denotes the low-rank mechanism that regulates the bottleneck, and the activation function used is ReLU. The embedding size is denoted by $h$, while $\alpha$ and $\beta$ represent the quantization numbers for the integer and fractional parts, respectively. Our access to special processors for variable quantization is limited, so we present a theoretical comparison of tunable parameters and inference speed in Table \ref{Table3-5}.

\begin{table*}[hbt!]
\caption{The efficiency of InA and other adaptation FT methods, in terms of trainable parameters, update speed (back-propagation) and inference (complexity)}
\label{Table3-5}
\centering

\begin{tabular}{cccc}
\toprule
Methods & \begin{tabular}[c]{@{}c@{}}Tunable Params\end{tabular}         & \begin{tabular}[c]{@{}c@{}}Inference\end{tabular}           & \begin{tabular}[c]{@{}c@{}}Update\end{tabular}   \\ \hline
\begin{tabular}[c]{@{}c@{}}Fully FT (float64)\end{tabular}    & \begin{tabular}[c]{@{}c@{}}$T1=3 \times L \times d^2 + d \times h$\end{tabular}              & $T1$                                                                                                                          &  $\mathcal{O}(n^{2})$, GeLU                                                               \\
\begin{tabular}[c]{@{}c@{}}Fully FT (Q$\alpha$i.$\beta$f)\end{tabular}   & \begin{tabular}[c]{@{}c@{}}$T2=3 \times L \times d^2 + d \times \frac{h}{\alpha+\beta}$\end{tabular}        & $T2$                                                                                                                                &  $\mathcal{O}(n^{2})$, GeLU                                                               \\
\begin{tabular}[c]{@{}c@{}}LoRA FT (float64)\end{tabular}   & \begin{tabular}[c]{@{}c@{}}$T3=2 \times L \times d \times r + d \times h$\end{tabular}          & $T1+T3$                                                                                                                             &  $\mathcal{O}(n)$, ReLU                                                               \\
\begin{tabular}[c]{@{}c@{}}LoRA FT  (Q$\alpha$i.$\beta$f)\end{tabular}   & \begin{tabular}[c]{@{}c@{}}$T4=2 \times L \times d \times r + d \times \frac{h}{\alpha+\beta}$\end{tabular}        & $T2+T4$                                                                                                                               &  $\mathcal{O}(n)$, ReLU                                                               \\ \bottomrule
\end{tabular}
\end{table*}

The Diffusion LM achieves competitive results when using the LoRA FT method on quantized embedding vectors. However, upon analyzing Table \ref{Table3-3} and Table \ref{Table3-4} in Appendix C, we observed a significant reduction of 90\% in trainable weights compared to the standard FT method. To assess the quality of the generated text, we evaluated DLMs' performance on different control tasks using $fluency$ (or \texttt{lm-score}) and \texttt{success} (or \texttt{ctrl}) metrics, as shown in Table \ref{Table3-1}. While our method did not lead to substantial improvements in these metrics (e.g., lower Loss and MSE or equivalent $fluency$ and \texttt{success} levels), it successfully reduced the trainable weights through the LoRA FT method. Furthermore, considering the combined factors of computing speed and inference complexity, LoRA fine-tuning method proved to be notably faster than full fine-tuning.

\subsection{Quantization Processing On Embedding Vectors}


The DLMs (Domain Language Models) are trained with various quantization methods and their Loss and Mean Squared Error (MSE) are compared before updating the weights under the control classifier. In Appendix C, Table \ref{Table3-3} and Table \ref{Table3-4} show the advantages of the quantization process for the generation LM based on BERT-base \cite{devlin2018bert}. It is observed that quantizing the integer part of the embedding vector leads to higher Loss and MSE, indicating weaknesses in methods that quantize the integer-part of embedding vectors during the fine-tuning of DLMs using pre-trained LMs. Figure \ref{Figure3-3} and Figure \ref{Figure3-6} in the Appendix illustrate the results obtained using different quantization methods, showing that the embedding space undergoes significant changes regardless of whether LoRA FT (LoRA Fine-Tuning) is employed or not. The reason behind this result is that LoRA tends to maintain the same embedding space as the classical fine-tuning method does.

\subsection{Controllable Text Generation Results}

\begin{table*}[t]
\caption{The QE-CDLM achieves a high success rate (\texttt{ctrl}, higher is better) and good fluency (\texttt{lm}, lower is better) across all five control tasks, outperforming the FUDGE and DLM baselines. }
\label{Table3-1}
\footnotesize\setlength\tabcolsep{2pt}
\centering
\begin{tabular}{p{0.09\linewidth}| p{0.06\linewidth}| p{0.056\linewidth} p{0.06\linewidth}| p{0.056\linewidth} p{0.06\linewidth}|  p{0.056\linewidth} p{0.06\linewidth}| p{0.056\linewidth} p{0.06\linewidth}| p{0.056\linewidth} p{0.06\linewidth}| p{0.15\linewidth}}
\toprule \multirow{2}{*}{\begin{tabular}[c]{@{}c@{}}Models\\ (DLM) \end{tabular}}
            & \multicolumn{1}{c|}{
            None} & \multicolumn{2}{c|}{Semantic} & \multicolumn{2}{c|}{POS} & \multicolumn{2}{c|}{Syntax Tree} & \multicolumn{2}{c|}{Syntax Spans} & \multicolumn{2}{c|}{Length}  & \multirow{2}{*}{\begin{tabular}[c]{@{}c@{}}Tunable Weights\\ \textbf{Basic}+Extra (\textbf{M})\end{tabular}}\\
                       & \texttt{lm}            & \texttt{ctrl}               & \texttt{lm}                & \texttt{ctrl}              & \texttt{lm}               & \texttt{ctrl}            & \texttt{lm}             & \texttt{ctrl}            & \texttt{lm}              & \texttt{ctrl}         & \texttt{lm}          \\ \hline 
\multicolumn{13}{c}{\textit{\textbf{SOTA Results}}}          \\ \hline
\begin{tabular}[c]{@{}c@{}}FUDGE \cite{yang2021fudge} \end{tabular}                 & -             & 69.9               & 2.83              & 27.0              & 7.96             & 17.9            & \textbf{3.39}           & 54.2            & 4.03            & 46.9         & 3.11       & - \\ \hline
\begin{tabular}[c]{@{}c@{}}DLM \cite{li2022diffusion} \end{tabular}                 & -             & \textbf{81.2}               & \textbf{2.55}              & \textbf{90.0}              & \textbf{5.16}             & \textbf{86.0}            & 3.71           & \textbf{93.8}            & \textbf{2.53}            & \textbf{99.9}         & \textbf{2.16}       & \textbf{86.0}+4.82 \\ \hline
\multicolumn{13}{c}{\textit{ \textbf{Fully/LoRA Fine-Tuning With Quantized Embedding Vectors}}}          \\ \hline
DLM-float64                     & 3.38/3.62              & 83.2/84.7                   & 3.62/3.43                  & 72.2/73.6                  & 5.45/5.50                 & 80.8/82.1                & 2.75/3.65               & 87.9/86.8                & 3.23/2.92                & 99.9/99.9             & 2.57/2.60            & (\textbf{86.0}+4.82)/(\textbf{0}+5.12) \\
DLM-Q0i.8f                   & 205/228              & \textbf{85.5}/85.0                   & \textbf{1.47}/1.65                  & 82.7/81.1                  & 3.22/3.44                 & \textbf{90.5}/90.0                & 1.77/1.57               & \textbf{94.8}/93.7                & \textbf{1.27}/1.34                & 99.9/99.9             & \textbf{2.29}/3.71            & (\textbf{86.0}+0.602)/(\textbf{0}+0.902) \\
DLM-Q0i.4f                   & 322/316              & 84.8/\textbf{85.9}                   & 1.64/\textbf{1.45}                  & 89.5/87.3                  & 3.80/5.72                 & 90.0/\textbf{90.1}                & \textbf{1.41}/\textbf{1.37}               & 93.7/\textbf{95.0}                & 1.35/1.40                & 99.9/99.9             & 2.74/3.59            & (\textbf{86.0}+0.301)/(\textbf{0}+0.601) \\
DLM-Ternary                   & 348/315              & 82.8/85.6                   & 1.99/1.48                  & 78.2/76.7                  & \textbf{2.38}/\textbf{1.86}                 & 88.0/90.0                & 3.27/1.64               & 85.2/94.7                & 2.40/\textbf{1.29}                & 99.9/99.9             & 2.87/2.64            & (\textbf{86.0}+0.151)/(\textbf{0}+0.451) \\
DLM-Binary                     & 336/235              & 84.7/82.2                   & 1.54/1.70                  & 80.5/78.7                  & 3.28/4.16                 & 89.0/86.4                & 3.17/3.05               & 91.9/90.5                & 1.46/1.57                & 99.9/99.9             & 2.86/\textbf{2.40}            & (\textbf{86.0}+0.075)/(\textbf{0}+0.375) \\
 \bottomrule
\end{tabular}
\end{table*}

Table \ref{Table3-1} summarizes the controllable text generation metrics (\texttt{ctrl} and \texttt{lm-score}).

\textbf{None.} Table \ref{Table3-1} showcases that the perplexity increases by orders of magnitude (up to 300 times) when the embedding vectors are quantized by 8-bit floating point format, 4-bit floating point format, ternary format, or binary format. The generated text without control is reported in \tablename~\ref{Table3-8} in the Appendix.

\textbf{Semantic.} Assume a $field$ (e.g., name) and $value$ (e.g., Vaults) is given, a sentence of $field=value$ is generated, and the success rate (\texttt{ctrl}) is obtained by calculating the exact match rate of ``$value$''. It is evident that quantized embedding vectors can be employed in the Standard and LoRA FT methods. An example of controlling the semantic value at 'Vaults' is illustrated in \tablename~\ref{Table3-16} in the Appendix.

\textbf{Parts-Of-Speech.}  Assume a sequence of parts-of-speech (POS) tags (including, e.g., ``PRON'', ``NOUN'', ``NOUN'', ``AUX'', and ``ADJ'') is given, a sequence of words whose POS tags (under an oracle POS tagger) match the target (e.g., Its customer rating is high) is generated. An example of controlling the POS of texts is depicted in Table \ref{Table3-13} in the Appendix. \figurename~\ref{Figure3-6} exhibits the 3D t-SNE map{---}with its 2D projected map{---}of the learned word embedding.

\textbf{Syntax Tree.} Assume a target syntactic parse tree is given as illustrated in Table \ref{Table3-2}), and the text whose syntactic parse matches the given parse is generated. The \texttt{success} is evaluated by parsing the generated text with an off-the-shelf parser \cite{kitaev2018constituency}, and F1 scores are calculated as well. \tablename~\ref{Table3-2} lists one syntactic parse tree example and emphasizes failing spans in \textcolor{red}{red} and \textbf{bold} for spans of interest. \tablename~\ref{Table3-14} in the Appendix depicts an example of controlling the Syntax Tree of texts.

\textbf{Syntax Spans.} Assume a target pair (span, syntactic category) is given, the text associated with a parse tree over the span [$i$; $j$] matches the target syntactic category (e.g., [$i=3$, $j=5$, prepositional phrase (PP)]) is generated. The \texttt{success} is quantified using the fraction of spans that exactly match. Table \ref{Table3-15} in the Appendix presents an example of controlling the Syntax Tree of texts.

\textbf{Length.} Assume a target with a fixed length (e.g., in $\{10, 15,\ldots, 40\}$), a sequence with a length within $\pm2$ of the target is generated. \tablename~\ref{Table3-12} in the Appendix presents an example of controlling the length of texts.

\begin{table*}[t]
\caption{Exemplify qualitative instances from the Syntax Tree control task using the associated quantization methods. The syntactic parse tree, characterized by nested brackets denoting constituents, is linearized, with adherence to standard PTB syntactic categories. There is an example of syntactic parse tree: (S (NP (NNP) (NNP)) (VP (VP (VBZ) (NP (JJ) (NN)) (\textit{\textbf{PP}} (IN \textit{\textbf{in}}) (\textit{\textbf{NP}} (DT \textit{\textbf{the}}) (JJ) (NN) (NN)))) (.) (NP (PRP)) (VP (VBZ) (NP (DT) (QP (CD) (IN) (IN) (CD)) (NN) (NN)))) (.) (CD)). The failing spans are colored \textcolor{red}{red}, and succeed spans are colored \textcolor{blue}{blue} in terms of the spans of interest (the \textbf{bold} text).}
\label{Table3-2}
\small
\centering
\begin{tabular}{p{0.09\linewidth} p{0.42\linewidth} p{0.42\linewidth}}
\toprule

\multicolumn{1}{c}{\begin{tabular}[c]{@{}c@{}}\textbf{Q-Methods}\\ \end{tabular}} & \multicolumn{1}{c}{\begin{tabular}[c]{@{}c@{}}\textit{\textbf{Fully FT With Quantized Embedding}}\end{tabular}}  & \multicolumn{1}{c}{\begin{tabular}[c]{@{}c@{}}\textit{\textbf{LoRA FT With Quantized Embedding}}\end{tabular}}           \\ \hline
\textbf{Diffusion-LM} \textbf{float64} & The Dumpling Tree is a restaurant \textcolor{red}{with} \textcolor{red}{a} cheap price range . It has a 5 out of 5 customer rating .                                                       & The Golden Curry is a pub near Café Brazil \textcolor{blue}{in} \textcolor{red}{\{\}} Riverside . It has a 3 out of 5 customer rating .                                                    \\ \hline
\textbf{Diffusion-LM} \textbf{Q0i.8f}  & Browns Cambridge serves English food \textcolor{blue}{in} \textcolor{blue}{the} high price range . It has a 1 out of 5 customer rating .                                                        & Browns Cambridge serves Indian food \textcolor{blue}{in} \textcolor{blue}{the} high price range . It achieves a 3 out of 5 customer rating .                                                  \\ \hline

\textbf{Diffusion-LM} \textbf{Q0i.4f}  & The Phoenix serves Chinese food \textcolor{blue}{in} \textcolor{red}{a} moderate price range . It has a 1 out of 5 customer rating .                                                        & The Rice Boat is a kid friendly Japanese restaurant \textcolor{blue}{in} \textcolor{red}{\{\}} Riverside . It has a 3 out of 5 customer rating .                                                   \\ \hline

\textbf{Diffusion-LM} \textbf{Ternay}  & Giraffe . It is kid friendly \textcolor{blue}{in} \textcolor{red}{a} moderate price range . It has a 3 out of 5 customer rating .                                                     & Strada serves Chinese food \textcolor{blue}{in} \textcolor{blue}{the} cheap price range . It has a 5 out of 5 customer rating .                                                   \\ \hline

\textbf{Diffusion-LM} \textbf{Binary}  & Bar . It provides Indian food \textcolor{red}{with} \textcolor{red}{a} cheap price range . It also a 5 out of 5 customer rating .                                                       & Aromi coffee shop serves Italian food \textcolor{red}{with} \textcolor{red}{a} cheap price range . It has a 5 out of 5 customer rating .                                                   \\ \bottomrule
\end{tabular}
\end{table*}

\section{Analysis and Discussion}

The empirical validation of quantized embedding vectors and LoRA FT on controllable Diffusion LMs is presented below. Two techniques are employed: (1) quantized vectors, which truncate partial information from original vectors, and (2) fine-tuning a pre-trained LM by reweighting pre-trained weights or tuning small tunable vectors (e.g., adaption vector). Three key questions are addressed and answered as follows: \textbf{Q1:} Does remodeling the embedding space by quantizing embedding vectors maintain or improve the quality of generated outputs? \textbf{Q2:} How can we quantize embedding vectors effectively to avoid negative effects? \textbf{Q3:} Can competitive performance be achieved using adapted FT methods instead of training DLMs from scratch, and what benefits can be gained from adapted FT methods?

\subsection{Should We Remodel the Task-Specific Embedding Space with Quantization Methods?}

The study investigates the impact of quantization methods on the embedding space. Instead of using the original word embeddings, we perform task-specific embeddings on tag words, resulting in a task-specific embedding space (e.g., controlling POS). Figure \ref{Figure3-6} shows that after quantizing the learned embedding vector using Q$0$i$8$f differs from the original baseline. Additionally, the metrics (fluency and success of generated texts in Table \ref{Table3-1} using quantization methods) demonstrate an overall improvement, indicating the advantage of quantizing the embedding vectors for task-specific control. The positive results suggest that quantization methods lead to positive changes in the embedding space.

Furthermore, the Syntax Tree example in Table \ref{Table3-2} also provides evidence of the positive impact of quantization on the embedding space. By employing the Q$0$i$8$f quantization method, the DLM under Standard FT shows improved stability in achieving success and higher fluency in generating proper spans at the corresponding position, which was challenging with the original embedding vector.

\subsection{How To Avoid a Negative Effect When Quantizing Embedding Vectors?}

The computation of embedding vectors is accelerated using vector quantization methods. However, the method that quantizes only the integer part of embedding vectors has a negative impact on DLMs, as shown in Table \ref{Table3-3} and Table \ref{Table3-4} of Appendix C. This quantization approach remodels the task-specific embedding space (e.g., control POS) and leads to larger quantized variables compared to those provided by the fractional part, resulting in a significant variation in the embedding space. As illustrated in Figure \ref{Figure3-3} of Appendix C, the embedding space generated by the Q$0$i$8$f quantization method outperforms that created by the Q$8$i$0$f method. These findings highlight the distinctiveness of fractional-part quantizing methods.

\subsection{What Can We Benefit from the Adaption FT Method?}

Finally, the focus of this study shifts to the impact of adaption fine-tuning, specifically its ability to reduce tunable weights while maintaining or even enhancing the quality of controllable Diffusion Language Models. In Appendix C, Table \ref{Table3-3} and Table \ref{Table3-4} present the results from two aspects: Loss and MSE. Although the LoRA FT method shows relatively lower performance, and in Table \ref{Table3-1}, LoRA FT exhibits lower metrics in terms of fluency and success, the differences between the two FT methods are minor. The significant improvement is that the tunable weights are notably reduced by over 90\%.

\section{Conclusion}

Quantized embedding vectors are proposed in DLM, a novel controllable diffusion language model. This approach offers several benefits: (1) it optimizes the task-specific embedding space, enabling faster convergence through quantization methods, (2) it reduces the perplexity of DLMs, and (3) it expedites the generation process with much less computation. However, there are several potential aspects to be improved over current research: (1) The decoding process remains relatively slow. One possible solution is to quantize and fine-tune the LM on large language datasets in the first step, which can significantly speed up the decoding process; (2) The intersection of multiple controls can be addressed by obtaining the sum of log-probabilities of several classifiers and using it as the gradient; (3) A more effective fine-tuning method should be developed to improve the performance of DLMs and thus reduce the tunable weights.








{\small
\bibliographystyle{IEEEtran}
\bibliography{main}
}




\newpage
\onecolumn
\appendix

\begin{table*}[ht]
\caption{Qualitative output of DLM without any control, where all the generated texts tried to exactly
match the target.}
\label{Table3-8}
\scriptsize
\centering
\begin{tabular}{p{0.1\linewidth} p{0.8\linewidth}}
\toprule
no target                                                   & without control-classifier                                                                                                                                                                                                                                                                                                                                                                                                                                                                                                                                   \\ \hline
\multirow{3}{*}{\begin{tabular}[c]{@{}l@{}}DLM-E2E\\ float64\\ Standard FT\end{tabular}} & START The Cambridge Blue is a French pub that is under £ 20 and is near the Café Brazil . \textbackslash{}n END \\ & START The Mill is a pub type Japanese food which has a price range under 20 pounds and is located on the riverside . \textbackslash{}n END \\ &  START There is a coffee shop near Café Brazil in city centre friendly restaurant offering English food with a price range of less than 20 . It is located in the center of the city near Express by Holiday Inn . The customer rating is low and the price range is less than £ 20 . \textbackslash{}n END                \\ \hline

\multirow{4}{*}{\begin{tabular}[c]{@{}l@{}}DLM-E2E\\ Q0i.8f\\ Standard FT\end{tabular}}  & START The Phoenix is an Italian restaurant by the river with located in the city centre. \textbackslash{}n END\\ & START There is a coffee shop in the riverside called The Olive Grove. It is not family - friendly, with a low customer rating, they can get English food. \textbackslash{}n END\\ & START One favorite Midsummer House offers high rated fast food, located near the Café Rouge.  END \textbackslash{}n END             \\ \hline

\multirow{3}{*}{\begin{tabular}[c]{@{}l@{}}DLM-E2E\\ Q0i.4f\\ Standard FT\end{tabular}}  & START The Cambridge Blue is a Italian pub that is below £ 30 and is near the Café Brazil . \textbackslash{}n END\\ & START Next to All Bar One is a low - priced restaurant called Green Man . They serve American food . \textbackslash{}n END \\ & START The Waterman has a price range of more than £ 30 located on the riverside serves Indian food . It has an excellent customer rating of 5 out of 5 and is known to no kids area . \textbackslash{}n END \\ \hline

\multirow{3}{*}{\begin{tabular}[c]{@{}l@{}}DLM-E2E\\ Ternary\\ Standard FT\end{tabular}}  & START Cocum is a coffee shop offering Japanese food . its customer rating is average with the UNK . \textbackslash{}n END\\ & START Cocum is a customer has price friendly restaurant . The customer rating is low . \textbackslash{}n END \\ & START Aromi is a pub located at City centre . its customer rating is 5 out of UNK . \textbackslash{}n END                                                                                                 \\ \hline
\multirow{3}{*}{\begin{tabular}[c]{@{}l@{}}DLM-E2E\\ Binary\\ Standard FT\end{tabular}}  & START Midsummer House , sushi restaurant and 5 star near All Bar One . \textbackslash{}n END \\ & START The Waterman , in the Riverside area , is not family - friendly , with a high customer rating and it serves cheap English food . \textbackslash{}n END \\ &  START Midsummer House provides Chinese food . It is near Café Rouge . Its customer rating is average . \textbackslash{}n END                \\ \hline  \hline
no target                                                   & without control-classifier   \\ \hline

\multirow{3}{*}{\begin{tabular}[c]{@{}l@{}}DLM-E2E\\ float64\\ LoRA FT\end{tabular}}   & START The Cambridge Blue is a Italian that costs more than £ 30 and is near the Café Brazil . \textbackslash{}n END \\ & START The Clowns fast food pub , has high prices and 3 of 5 star rating . It is located by the riverside . \textbackslash{}n END \\ & START There is a coffee shop near Café Brazil in city centre
Blue is an English pub with a price range of over 30 near Café Brazil . \textbackslash{}n END               \\ \hline

\multirow{3}{*}{\begin{tabular}[c]{@{}l@{}}DLM-E2E\\ Q0i.8f\\ LoRA FT\end{tabular}}    & START The Rice Boat is a high priced restaurant located in Riverside near Express by Holiday Inn UNK Indian food . They have an average customer rating of 3 out of 5 . They are in the city centre .  \textbackslash{}n END\\ & START The Olive Grove is a pub , with a moderate price range . It has English food and is just UNK , it is in the riverside area not recommended for families with small children . \textbackslash{}n END\\ & START A coffee shop , Cocum , with a customer rating of 1 out of 5 . It has a superb view of 4 , and is not family - friendly in the moderate price range . \textbackslash{}n END\             \\ \hline

\multirow{3}{*}{\begin{tabular}[c]{@{}l@{}}DLM-E2E\\ Q0i.4f\\ LoRA FT\end{tabular}}    & START Zizzi , has a customer rating of 3 out of 5 , is kid friendly and serves French food . \textbackslash{}n END\\ & START The Waterman is a family friendly restaurant providing fast food in a high price range for Indian food . It has an average customer rating of 3 out of 5 and is located in the city centre . \textbackslash{}n END \\ & START The Vaults is a restaurant that serves Japanese food and is child friendly . It is low priced . \textbackslash{}n END   \\ \hline

\multirow{3}{*}{\begin{tabular}[c]{@{}l@{}}DLM-E2E\\ Ternary\\ LoRA FT\end{tabular}}    & START Browns Cambridge serves Japanese food . It is located in the city centre near the The Sorrento . \textbackslash{}n END\\ & START Zizzi , has a customer rating of 3 out of 5 , is kid friendly and serves French food . \textbackslash{}n END \\ & START The Green Man is a food restaurant with a moderate price range . It has English food and is children friendly , it is in the riverside area , near All Bar One in Riverside . \textbackslash{}n END                                                                                              \\ \hline
\multirow{3}{*}{\begin{tabular}[c]{@{}l@{}}DLM-E2E\\ Binary\\ LoRA FT\end{tabular}}     & START The Fitzbillies has a customer rating of 3 out of 5 , is kid friendly and serve Italian food . \textbackslash{}n END\\ & START The Waterman is a low priced restaurant located in the riverside that mid price range . It is child friendly . \textbackslash{}n END \\ & START Strada is a cheap restaurant close to the Rainbow Vegetarian Café . It . \textbackslash{}n END                 \\ \bottomrule
\end{tabular}
\end{table*}

\begin{table*}[ht]
\caption{Randomly sampled examples generated by unconditional sampling Diffusion-LM trained
on ROCStories.}
\label{Table3-9}
\scriptsize
\centering
\begin{tabular}{p{0.1\linewidth} p{0.8\linewidth}}
\toprule
\multirow{3}{*}{\begin{tabular}[c]{@{}l@{}}ROCStories\\ float64\\ Standard FT\end{tabular}} & START I wanted to see an UNK . She asked my sister to UNK in a band . I agreed with it . So she went to the store . But it was hard to play again . END                                        \\
& START Allie wanted to go back to the park pool . She was going to go to the park too fast . She decided to find some money instead . She ended up getting down . But then she went . END       \\
& START I had an old UNK UNK . The UNK was against a very big UNK . It was very UNK and friendly . I tried to UNK out good to fix the UNK . It was well in the UNK UNK . END                     \\ \cline{2-2} 
\multirow{3}{*}{\begin{tabular}[c]{@{}l@{}}ROCStories\\ Q0i.8f\\ Standard FT\end{tabular}}  & START UNK wanted to be an UNK . She was confident she would be a UNK test . She studied all of her students . She was so great . She could n't wait to play again . END          \\
& START I had an assignment in UNK . The UNK tasted like a very large person cost . I had UNK and ok . I decided to UNK out the UNK into the UNK . I put it in the week later . END                                    \\
& START The man was outside for a woman . The woman found some UNK . The man complained to the car . The woman fixed the UNK . The UNK left . END                \\ \cline{2-2} 
\multirow{3}{*}{\begin{tabular}[c]{@{}l@{}}ROCStories\\ Q0i.4f\\ Standard FT\end{tabular}}  & START UNK wanted to be an UNK . Her wife thought she would be a UNK test . She got all of her eye . She was UNK to . But it was high , she sighed . END                                 \\
& START Jamie needed to go to the on bar one day . She decided to go to a UNK the night . She decided to go . They went to the bar . It was nice time they went outside . END                                           \\
& START Tom was a UNK . He decided to cook one from his friends . His friends tried to UNK his UNK . It kept to read UNK . Tom decided to let the rest of his friends on him . END                                            \\ \cline{2-2} 
\multirow{3}{*}{\begin{tabular}[c]{@{}l@{}}ROCStories\\ Ternary\\ Standard FT\end{tabular}} & START UNK wanted to be an UNK . She asked her friends to UNK the next month . They played UNK for her . UNK won the UNK UNK . All her kids wanted to play again . END                          \\
& START Jamie and his friends went to the kitchen to make some UNK . They went to check the fire . Once they were UNK up . They tried to decide what to do . Finally , they helped UNK . END     \\
& START The man had a UNK . He asked his wife for some UNK . The man refused . His wife UNK the man at the UNK . The man laughed . END                                                           \\ \cline{2-2} 
\multirow{3}{*}{\begin{tabular}[c]{@{}l@{}}ROCStories\\ Binary\\ Standard FT\end{tabular}}  & START UNK wanted to fly an UNK . She had thought she would be a UNK test . She got out of her car . She was UNK to . All her daughter watch a play again . END                                 \\
& START I decided to go back to the mall at store in UNK . I did n't get UNK very fast . I decided to stay in and tried to go for a UNK . She said to try UNK . I was very sad . END             \\
& START It was UNK 's birthday night . UNK was UNK UNK . He called and UNK went to his friends . He was at the UNK . They UNK himself . END                                                      \\ \cline{2-2} 
\multirow{3}{*}{\begin{tabular}[c]{@{}l@{}}ROCStories\\ float64\\ LoRA FT\end{tabular}}   & START I wanted to play UNK in a UNK UNK UNK . I went to the bank . The UNK was UNK and UNK on the UNK . The UNK went and cut all of the UNK . I decided to make the UNK at home . END          \\
& START Suzy wanted to go to the party . So she needed some UNK . So she wanted to go . So she was at the UNK . She loved it . END                                                               \\
& START Ann wanted to quit UNK . She bought UNK for a month . The store was not very difficult . Unfortunately , she was smoking and UNK . She had to hold the UNK out of her UNK . END          \\ \cline{2-2} 
\multirow{3}{*}{\begin{tabular}[c]{@{}l@{}}ROCStories\\ Q0i.8f\\ LoRA FT\end{tabular}}    & START Joe was in the UNK grade . He was still in UNK . He was on by the UNK . Eventually he fell through the UNK with his bag . Joe came home and fell asleep . END                                                                                                                                                                                               \\
& START Today UNK took the trip to the UNK . She was great around UNK UNK . She picked up by the UNK and UNK . She enjoyed the view . Afterwards she got the UNK . END                                                                                                                                                                                                \\
& START UNK was an UNK UNK . She knew she wanted to learn a new idea . She quit her UNK 's class and listened to her school . She hit a UNK friend . When she got home , she was thrilled with her older and UNK story . END                                                                                                                                                                                                \\ \cline{2-2} 
\multirow{3}{*}{\begin{tabular}[c]{@{}l@{}}ROCStories\\ Q0i.4f\\ LoRA FT\end{tabular}}    & START Tom wanted to play an UNK . His wife 's family would UNK for a month . Tom 's UNK became UNK . Tom called the UNK to play . Tom was excited to play again . END                                                                                                                                                                                               \\
& START The man was UNK . A stranger asked his money for a UNK . The stranger listened to the man . The stranger fixed the UNK . The man left . END                                                                                                                                                                                               \\
& START Tom was out to a UNK bar . It UNK from a friend . His friends decided to UNK over UNK . Tom had to take a friend for a bit of the fight . Tom and his friend went drunk . END                                                                                                                                                                                               \\ \cline{2-2} 
\multirow{3}{*}{\begin{tabular}[c]{@{}l@{}}ROCStories\\ Binary\\ LoRA FT\end{tabular}}    & START Tom wanted to be a UNK UNK . It thought it would be a little . He decided to sign up the local house to join UNK . Tom decided to leave the fight . He and his coworkers went fine . END \\
& START Tom was on a vacation . He got a new job . He was really excited to go away . Hours later , Tom 's girlfriend was UNK and UNK . It was almost an UNK UNK . END                           \\
& START Tom was walking to the party . There was a loud it would take left . Tom threw it out of the house to Tom . Tom tried to take the party . His friend tried to take him away for it . END \\ \cline{2-2} 
\multirow{3}{*}{\begin{tabular}[c]{@{}l@{}}ROCStories\\ Ternary\\ LoRA FT\end{tabular}}   & START UNK wanted to be an UNK . She told her she would write a metal UNK . She got out of her creek . She was UNK to UNK . The doctor made a perfect UNK . END                                 \\
& START A boy went to UNK in UNK . There was a UNK in the UNK UNK . The teacher liked it . He got the UNK . He moved UNK . END                                                                   \\
& START The man had 's for the UNK . The UNK was their UNK . The man listened to the UNK . He yelled at the UNK . The man threw the new UNK . END                                                \\ \bottomrule
\end{tabular}
\end{table*}

\begin{table*}[ht]
\caption{Randomly sampled examples generated by unconditional sampling Diffusion-LM trained on WikiText2.}
\label{Table3-10}
\scriptsize
\centering
\begin{tabular}{p{0.1\linewidth} p{0.8\linewidth}}
\toprule
WikiText2 float64 Standard FT & START Stanley UNK Green ( 22 February 1915 – 4 December 1993 ) , known as the UNK Man , was a human UNK who became a well @-@ known figure in central London in the latter half of the 20th century . 
  Green UNK Oxford Street in the West End for 25 years , from 1968 until 1993 , with a UNK                     \\ \hline 
WikiText2 Q0i.8f Standard FT & START Perfect Dark is a UNK release of the first @-@ person UNK video game by the same name . UNK by < unk > Studios and published by UNK Game Studios a decade after the original 's 2000 release , the UNK features several UNK UNK , including higher UNK UNK and UNK , a higher frame rate , and a UNK UNK          \\  \hline 
WikiText2 Q0i.4f Standard FT  & START UNK Athletic Football Club is a professional association football club based in the town of UNK , Kent , England . The club was formed in 1983 after the UNK of the town 's previous club , UNK UNK , whose place in the Southern League was taken by the new club . In the 1989 – 90 season UNK Athletic won the                                 \\ \hline 
WikiText2 Ternary Standard FT & START Central Area Command was one of several UNK based UNK raised by the Royal Australian Air Force ( RAAF ) during World War II . It was formed in March 1940 , and covered the central portion of New South Wales . UNK at UNK , Central Area Command was primarily responsible for air defence , UNK reconnaissance and UNK of the sea                          \\ \hline 
WikiText2 Binary Standard FT  & START The Rocky Mountain Horse is a UNK breed developed in the state of UNK in the United States . Despite its name , it originated not in the Rocky UNK , but instead in the UNK UNK . A foundation UNK , brought from the western United States to eastern UNK around 1890 , began the Rocky Mountain type in the late 19th                                 \\ \hline 
WikiText2 float64 LoRA FT   & START The Battle of UNK was an UNK in the UNK campaign of the American UNK War fought in the village of UNK , Vermont . Vermont was then a UNK territory sometimes called the New UNK UNK , claimed by New York , New UNK , and the newly organized and not yet recognized but de UNK independent government of Vermont . On          \\ \hline
WikiText2 Q0i.8f LoRA FT    & START Le UNK de UNK was a political UNK written by UNK UNK in UNK . With the French Revolution into its fourth year , civil war had spread across France between various rival political UNK . UNK was involved in military action , on the government 's side , against some UNK cities of southern France . It was during these events ,         \\ \hline 
WikiText2 Q0i.4f LoRA FT    & START UNK Tropical Storm UNK in 1984 caused 100 year UNK in South Africa and record rainfall in UNK . The fourth named storm of the season , UNK developed on January 16 off the northeast coast of UNK . With a ridge to the north , the storm tracked generally westward and later UNK . On January 21 , UNK struck eastern UNK        \\ \hline
WikiText2 Ternary LoRA FT    & START UNK is a station on the UNK Line ( line 6 ) of the UNK UNK in UNK . UNK between UNK UNK and UNK stations , it is the first station after the UNK Line leaves the UNK Line . The station is located 6 @.@ 1 kilometres ( 3 @.@ 8 mi ) from < unk > station . UNK is \\ \hline 
WikiText2 Binnary LoRA FT  & START UNK spacing is the UNK space between UNK in UNK text . It is a matter of UNK UNK . Since the introduction of UNK @-@ type UNK in Europe , various sentence spacing UNK have been used in languages with a Latin UNK . These include a normal word space ( as between the words in a sentence ) , a single                                                \\ \bottomrule
\end{tabular}
\end{table*}

\begin{table*}[ht]
\caption{Randomly sampled examples generated by unconditional sampling Diffusion-LM trained on WikiText103.}
\label{Table3-11}
\scriptsize
\centering
\begin{tabular}{p{0.1\linewidth} p{0.8\linewidth}}
\toprule
WikiText103 float64 Standard FT & START UNK was an ironclad ironclad ship built by the Italian Italian Regia Marina in the late 1890s . The War , was the lead ship of the first she was laid laid down in and the Battle of and national . First the of , in part which April was 1914 in the . was 1914 was in with the in the Mediterranean                     \\ \hline 
WikiText103 Q0i.8f Standard FT & START Typhoon UNK , known in the Philippines as Typhoon UNK , was the the first named typhoon of the history , the due to the most Pacific in the region of . Japan from west to August in the the state California , , Philippines the moved to , , August 16 as , a not to generally UNK southern the in .          \\  \hline 
WikiText103 Q0i.4f Standard FT  & START " UNK UNK " is the seventh episode of the second season of the American comedy television series The Office . The episode was written and directed by UNK , on January 12 , 2008 , and directed by Ryan , and UNK . 
  on UNK episode The the , , at American series on May NBC , 2008 of , and                                 \\ \hline 
WikiText103 Ternary Standard FT & START The UNK @-@ UNK was a British of the first unit of the Croatian during the late War II . The village of UNK the UNK of , the the in , the and was during in the of Battle the city the of Army the in , the of a over and the , last of . during , UNK the was                          \\ \hline 
WikiText103 Binary Standard FT  & START " The UNK " is the fifth episode of the ninth season of the American comedy comedy television series The Office , and UNK episode and the episode overall of . originally It on on , in aired NBC States November , on 2007 episode and , directed Paul United was . directed @-@ and UNK The and written UNK Mark by by                                 \\ \hline 
WikiText103 float64 LoRA FT   & START " UNK UNK " is a song by American recording recording artist Beyoncé artist 's UNK from fifth studio album Dream ( 2013 ) . Originally released on 16 , 1993 , 10 , EP , her and second album , the Japanese . The by was written The song , performed by American and producer , Gaga , The and , composed          \\ \hline
WikiText103 Q0i.8f LoRA FT    & START The UNK Creek was a 10 @.@ feet ( 6 @.@ km ) flows in UNK , from Croatia , the UNK of the Columbia River in the flows on the River Thames , in the United States , Canada in , and Mexico United the Kingdom , of the and the . It , flows the along the of River UNK has          \\ \hline 
WikiText103 Q0i.4f LoRA FT    & START The Calgary UNK is an English professional association football between the English team and the and population of city the , , York in , , the and home , UNK the 
  . UNK the in Atlantic of the The in team of . , the and @.@ in , 9 – 2 previous 3 25 . a km the ( )        \\ \hline
WikiText103 Ternary LoRA FT    & START UNK Arthur " UNK " UNK ( born 29 September 1982 ) is an Irish actor of UNK for UNK , who , among small books , and UNK UNK . She was one of UNK 's UNK first films , in the British Empire from the 1930s , , 2011 , in 1997 and Australia . The films of a career in \\ \hline 
WikiText103 Binary LoRA FT  & START SMS UNK was the second of the Königsberg class class of battleships of Navy the her ship lead , named , and her ) shipyard , the in down was in ( ship , and commissioned sister the was launched 1906 for . , in in and , She 1940 launched in . September the ship was in and the the in last                                 \\ \bottomrule
\end{tabular}
\end{table*}

\begin{table*}[]
\caption{Qualitative output of the length control tasks, where all the generated texts tried to exactly
match the target length. We mark the words exceeding the target length red.}
\label{Table3-12}
\scriptsize
\centering
\begin{tabular}{ll}
\toprule
target length                                                   & 10                                                                                                                                                                                                                                                                                                                                                                                                                                                                                                                                   \\ \hline
\multirow{3}{*}{\begin{tabular}[c]{@{}l@{}}DLM-E2E\\ float64\\ Standard FT\end{tabular}} & START The Vaults is a Japanese restaurant . \textbackslash{}n END \\ & START The Waterman is not family friendly . \textbackslash{}n END \\ & START The Eagle serves expensive Indian food . \textbackslash{}n END                 \\ \hline

\multirow{3}{*}{\begin{tabular}[c]{@{}l@{}}DLM-E2E\\ Q0i.8f\\ Standard FT\end{tabular}}  & START The Punter is an Italian restaurant . \textbackslash{}n END\\ & START The Punter is an Indian restaurant . \textbackslash{}n \textbackslash{}n END\\ & START The Vaults is a French restaurant . \textbackslash{}n END            \\ \hline

\multirow{3}{*}{\begin{tabular}[c]{@{}l@{}}DLM-E2E\\ Q0i.4f\\ Standard FT\end{tabular}}  & START The Waterman is a Chinese restaurant . \textbackslash{}n END\\ & START The Twenty Two serves French food . \textbackslash{}n END \\ & START The Twenty Two serves Indian food . \textbackslash{}n END \\ \hline

\multirow{3}{*}{\begin{tabular}[c]{@{}l@{}}DLM-E2E\\ Ternary\\ Standard FT\end{tabular}} & START The Waterman is an Indian restaurant . \textbackslash{}n END\\ & START The Twenty Two serves French food . \textbackslash{}n END \\ & START The Waterman is not family friendly . \textbackslash{}n END                                                                                                \\ \hline

\multirow{3}{*}{\begin{tabular}[c]{@{}l@{}}DLM-E2E\\ Binary\\ Standard FT\end{tabular}}  & START The Twenty Two serves Chinese food . \textbackslash{}n END\\ & START The Vaults is a Chinese restaurant . \textbackslash{}n END \\ & START Wildwood is a family friendly restaurant . \textbackslash{}n END                 \\ \hline  \hline
target length                                                   & 15                                                                                                                                                                                                                                                                                                                                                                                                                                                                                                                                   \\ \hline
\multirow{3}{*}{\begin{tabular}[c]{@{}l@{}}DLM-E2E\\ float64\\ LoRA FT\end{tabular}}   & START The Eagle is a French restaurant with a low customer rating . \textbackslash{}n END \\ & START The Eagle is a Chinese restaurant with a low customer rating . \textbackslash{}n END \\ & START The Cambridge Blue is a Chinese restaurant with a high rating . \textbackslash{}n END                \\ \hline

\multirow{3}{*}{\begin{tabular}[c]{@{}l@{}}DLM-E2E\\ Q0i.8f\\ LoRA FT\end{tabular}}    & START The Dumpling Tree is an English food restaurant with low prices . \textbackslash{}n END\\ & START The Eagle is a Chinese restaurant with a low customer rating . \textbackslash{}n END\\ & START The Vaults is a family friendly restaurant that serves English food . \textbackslash{}n END             \\ \hline

\multirow{3}{*}{\begin{tabular}[c]{@{}l@{}}DLM-E2E\\ Q0i.4f\\ LoRA FT\end{tabular}}    & START The Cambridge Blue is a Chinese restaurant with high customer ratings . \textbackslash{}n END\\ & START The Dumpling Tree is a restaurant with a cheap price range . \textbackslash{}n END \\ & START The Cambridge Blue is an Italian restaurant with high customer ratings . \textbackslash{}n END    \\ \hline

\multirow{3}{*}{\begin{tabular}[c]{@{}l@{}}DLM-E2E\\ Ternary\\ LoRA FT\end{tabular}}   & START The Cambridge Blue is a Chinese restaurant with high customer ratings . \textbackslash{}n END\\ & START Loch Fyne is a family friendly restaurant that serves English food . \textbackslash{}n END \\ & START The Eagle is a fast food with a low customer rating . \textbackslash{}n END                                                                                               \\ \hline

\multirow{3}{*}{\begin{tabular}[c]{@{}l@{}}DLM-E2E\\ Binary\\ LoRA FT\end{tabular}}    & START The Waterman is a family friendly restaurant located in the riverside . \textbackslash{}n END\\ & START The Eagle is a family restaurant located in the city centre . \textbackslash{}n END \\ & START The Mill is an Italian food restaurant in the city centre . \textbackslash{}n END                  \\ \bottomrule
\end{tabular}
\end{table*}

\begin{table*}[ht]
\caption{Qualitative output of the POS control tasks. The target POS is the sequence of gold parts-of-speech tags that the generated texts should match.}
\label{Table3-13}
\scriptsize
\centering
\begin{tabular}{p{0.1\linewidth} p{0.8\linewidth}}
\toprule
target pos                                                   & ('START', 'PROPN', 'AUX', 'DET', 'NOUN', 'VERB', 'NOUN', 'ADJ', 'NOUN', 'PUNCT', 'PRON', 'NOUN', 'NOUN', 'AUX', 'ADJ', 'ADP', 'END')                                                                                                                                                                                                                                                                                                                                                                                                                                                                                                                                   \\ \hline
\multirow{3}{*}{\begin{tabular}[c]{@{}l@{}}DLM-E2E\\ float64\\ Standard FT\end{tabular}} & START Zizzi is a pub serving river Japanese food . It has a customer rating of 1 out of 5 and is child friendly . \textbackslash{}n END \\ & START Wildwood is a pub that serves Indian food . its price range is high . \textbackslash{}n END \\ & START Zizzi is a pub that serves Indian food . Its customer rating is high . \textbackslash{}n END                \\ \hline

\multirow{3}{*}{\begin{tabular}[c]{@{}l@{}}DLM-E2E\\ Q0i.8f\\ Standard FT\end{tabular}}  & START Clowns is a pub located family Italian riverside. Its customer rating is high \textbackslash{}n END\\ & START Cocum is a restaurant has restaurant average rating. It customer rating is average \textbackslash{}n \textbackslash{}n END\\ & START Strada is a price located family friendly restaurant. customer rating is average \textbackslash{}n END             \\ \hline 

\multirow{3}{*}{\begin{tabular}[c]{@{}l@{}}DLM-E2E\\ Q0i.4f\\ Standard FT\end{tabular}}  & START Fitzbillies is a coffee has customer Italian food . It customer rating is high \textbackslash{}n END\\ & START Strada is a price has family friendly restaurant . its customer rating is high . \textbackslash{}n END \\ & START Strada is a low - rating Italian restaurant . It is family is friendly \textbackslash{}n END \\ \hline

\multirow{3}{*}{\begin{tabular}[c]{@{}l@{}}DLM-E2E\\ Ternary\\ Standard FT\end{tabular}}  & START Cocum is a coffee shop offering Japanese food . its customer rating is average with the UNK . \textbackslash{}n END\\ & START Cocum is a customer has price friendly restaurant . The customer rating is low . \textbackslash{}n END \\ & START Aromi is a pub located at City centre . its customer rating is 5 out of UNK . \textbackslash{}n END                                                                                                 \\ \hline
\multirow{3}{*}{\begin{tabular}[c]{@{}l@{}}DLM-E2E\\ Binary\\ Standard FT\end{tabular}}  & START Zizzi is a pub that offers Italian food . its customer rating is high in it is not kids friendly . \textbackslash{}n END\\ & START Zizzi is a family - friendly Italian pub . The customer rating is low of END \textbackslash{}n END \\ & START Giraffe is a family located restaurant serving pub . Its French food is French in the UNK . \textbackslash{}n END                \\ \hline  \hline
target pos                                                   & ('START', 'PROPN', 'AUX', 'DET', 'NOUN', 'VERB', 'NOUN', 'ADJ', 'NOUN', 'PUNCT', 'PRON', 'NOUN', 'NOUN', 'AUX', 'ADJ', 'ADP', 'END')  \\ \hline

\multirow{3}{*}{\begin{tabular}[c]{@{}l@{}}DLM-E2E\\ float64\\ LoRA FT\end{tabular}}   & START Clowns is a pub in the city centre . Its customer rating is low with price range is more than £ 30 . \textbackslash{}n END \\ & START Cocum is a restaurant with a high rating . It is family - friendly . \textbackslash{}n END \\ & START Cocum is a family friendly Japanese coffee shop . It customer rating is low with prices are below average . \textbackslash{}n END               \\ \hline

\multirow{3}{*}{\begin{tabular}[c]{@{}l@{}}DLM-E2E\\ Q0i.8f\\ LoRA FT\end{tabular}}    & START Zizzi is a family - - friendly restaurant . This coffee shop is located in the City are are are your are are to are are new are are are UNK are when are are are . \textbackslash{}n END\\ & The Waterman is a pub located on the riverside . It customer rating is low \textbackslash{}n END\\ & START Strada is a centre located family friendly restaurant . It is customer rated friendly \textbackslash{}n END\             \\ \hline

\multirow{3}{*}{\begin{tabular}[c]{@{}l@{}}DLM-E2E\\ Q0i.4f\\ LoRA FT\end{tabular}}    & START Zizzi is a pub that provides Italian food . It customer rating is low \textbackslash{}n END\\ & START Strada is a customer offers child friendly restaurant .  customer rating is low \textbackslash{}n END \\ & START Zizzi is a pub located food Italian food . It city family is friendly \textbackslash{}n END   \\ \hline

\multirow{3}{*}{\begin{tabular}[c]{@{}l@{}}DLM-E2E\\ Ternary\\ LoRA FT\end{tabular}}    & START Zizzi is a pub providing UNK Italian food . It customer rating is high . \textbackslash{}n END\\ & START Wildwood is a coffee located serving Italian food . The customer rating is high in UNK . \textbackslash{}n END \\ & START Cocum is a restaurant has a high rating . It is family - friendly with UNK . \textbackslash{}n END                                                                                              \\ \hline
\multirow{3}{*}{\begin{tabular}[c]{@{}l@{}}DLM-E2E\\ Binary\\ LoRA FT\end{tabular}}     & START The Waterman is a family friendly restaurant located in the riverside . \textbackslash{}n END\\ & START Wildwood is a pub has a low rating . The price range is high with END \textbackslash{}n END \\ & START Fitzbillies is a moderately priced family friendly restaurant . It customer rating is 1 out of 5 . It is child friendly . \textbackslash{}n END                 \\ \bottomrule
\end{tabular}
\end{table*}

\begin{table*}[ht]
\caption{Exemplify qualitative instances from the Syntax Tree control task using the associated quantization methods. The syntactic parse tree, characterized by nested brackets denoting constituents, is linearized, with adherence to standard PTB syntactic categories. There is an example of syntactic parse tree: (TOP (S (ADVP (RB Next) (PP (IN to) (NP (DT the) (NNP Yippee) (NNP Noodle) (NNP Bar)))) (NP (EX there)) (VP (VBZ is) (NP (NP (DT a) (ADJP (RB moderately) (JJ good)) (NN restaurant)) (VP (VBN called) (S (NP (NNP Alimentum)))))) (. .).}
\label{Table3-14}
\scriptsize
\centering
\begin{tabular}{p{0.1\linewidth} p{0.8\linewidth}}
\toprule
target tree                                                   & (TOP (S (ADVP (RB Next) (PP (IN to) (NP (DT the) (NNP Yippee) (NNP Noodle) (NNP Bar)))) (NP (EX there)) (VP (VBZ is) (NP (NP (DT a) (ADJP (RB moderately) (JJ good)) (NN restaurant)) (VP (VBN called) (S (NP (NNP Alimentum)))))) (. .) (. \textbackslash{}n )))                                                                                                                                                                                                                                                                                                                                                                                                                                                                                                                                  \\ \hline
\multirow{3}{*}{\begin{tabular}[c]{@{}l@{}}DLM-E2E\\ float64\\ Standard FT\end{tabular}} &  \textbackslash{}n END START Near Café Rouge there is a fast food place named Cotto .\\ & \textbackslash{}n END START In the city centre there is a moderately priced restaurant called Alimentum . \\ & \textbackslash{}n END START In the city centre there is a family friendly restaurant called Alimentum .                  \\ \hline

\multirow{3}{*}{\begin{tabular}[c]{@{}l@{}}DLM-E2E\\ Q0i.8f\\ Standard FT\end{tabular}}  & Burger King in the city centre . It also a low customer rating quality \textbackslash{}n END START\\ & Japanese food in the riverside area . It also a your friendly place quality Giraffe . \textbackslash{}n END START\\ & coffee shop in the city centre area that also a high price range quality No . \textbackslash{}n END START            \\ \hline

\multirow{3}{*}{\begin{tabular}[c]{@{}l@{}}DLM-E2E\\ Q0i.4f\\ Standard FT\end{tabular}}  & START UNK of the outskirts cheaper city there is a low priced pub named Wildwood . \textbackslash{}n END\\ & UNK say with a low price range there is a family friendly place called Cocum . \textbackslash{}n END START \\ & START Located in the city centre area there is a kid friendly pub called The Olive Grove which serves food at a moderate price range . \textbackslash{}n END \\ \hline

\multirow{3}{*}{\begin{tabular}[c]{@{}l@{}}DLM-E2E\\ Ternary\\ Standard FT\end{tabular}}  & UNK part near the river , The inexpensive is a family friendly restaurant called Zizzi . \textbackslash{}n END START\\ & . \textbackslash{}n END START By of Rouge there is a family friendly pub called Zizzi  \\ & START Located near The Bakers , what it is a family friendly restaurant called Zizzi . \textbackslash{}n END                                                                                                 \\ \hline
\multirow{3}{*}{\begin{tabular}[c]{@{}l@{}}DLM-E2E\\ Binary\\ Standard FT\end{tabular}}  & START Taste of Cambridge , Indian on the is the family friendly and is high . \textbackslash{}n END\\ & Indian There is a rather Indian Green Man is a family friendly and is situated at the riverside . \textbackslash{}n END START\\ & Indian quality food . \textbackslash{}n END START Strada is a family friendly restaurant located on Indian .                 \\ \hline  \hline

target tree                                                   & (TOP (S (ADVP (RB Next) (PP (IN to) (NP (DT the) (NNP Yippee) (NNP Noodle) (NNP Bar)))) (NP (EX there)) (VP (VBZ is) (NP (NP (DT a) (ADJP (RB moderately) (JJ good)) (NN restaurant)) (VP (VBN called) (S (NP (NNP Alimentum)))))) (. .) (. \textbackslash{}n ))) \\ \hline

\multirow{3}{*}{\begin{tabular}[c]{@{}l@{}}DLM-E2E\\ float64\\ LoRA FT\end{tabular}}   & English food in the higher price range there is a family friendly pub called Cocum . \textbackslash{}n END START \\ & Indian food at a high price . It is a family friendly restaurant called The Vaults . \textbackslash{}n END START\\ & English food in the higher price range there is a family friendly pub called Cocum . \textbackslash{}n END START              \\ \hline

\multirow{3}{*}{\begin{tabular}[c]{@{}l@{}}DLM-E2E\\ Q0i.8f\\ LoRA FT\end{tabular}}    & is located near the Café Rouge . It is a family friendly pub called Zizzi . \textbackslash{}n END START\\ & START When in riverside Near Café Rouge there is a family friendly restaurant called The Golden Curry . It has a low customer rating . \textbackslash{}n END\\ & English food in the high price range and has a customer rating of 1 out of 5 . \textbackslash{}n END START             \\ \hline

\multirow{3}{*}{\begin{tabular}[c]{@{}l@{}}DLM-E2E\\ Q0i.4f\\ LoRA FT\end{tabular}}    & city centre near Café Rouge . However it is not family friendly , is Strada . \textbackslash{}n END START\\ & START Next to The Bakers , twenty there is a family friendly restaurant called The Golden Curry , are got a low customer rating . \textbackslash{}n END \\ & Indian food in the city centre are It is not kid friendly and moderately priced . \textbackslash{}n END START  \\ \hline

\multirow{3}{*}{\begin{tabular}[c]{@{}l@{}}DLM-E2E\\ Ternary\\ LoRA FT\end{tabular}}    & START Right near The Six Bells , it is a family friendly restaurant called Fitzbillies . \textbackslash{}n END\\ & START Right near Rainbow Vegetarian Café , Strada is a family friendly restaurant with wonderful UNK , fruit UNK cheeses , wines and fruit . \textbackslash{}n END \\ & Indian food in the riverside area . It is a family friendly pub called Giraffe . \textbackslash{}n END START                                                                                             \\ \hline
\multirow{3}{*}{\begin{tabular}[c]{@{}l@{}}DLM-E2E\\ Binary\\ LoRA FT\end{tabular}}     & START There is a fast Indian area restaurant with a family friendly and is on the Ranch . \textbackslash{}n END\\ & Indian coffee shop . \textbackslash{}n END START It is a family friendly restaurant called Strada . \\ &  \textbackslash{}n END START There is bite area restaurant is a family friendly restaurant called The Vaults , . It it is not family - friendly .                 \\ \bottomrule
\end{tabular}
\end{table*}

\begin{table*}[ht]
\caption{The qualitative results from the syntax span control tasks are presented. The target span [$i$; $j$; $label$] indicates that the span from position $i$ to position $j$ is expected to be a constituent with a specified label. Failed spans are highlighted in \textcolor{red}{red}, while correct spans are color-coded in \textcolor{blue}{blue}.}
\label{Table3-15}
\scriptsize
\centering
\begin{tabular}{p{0.1\linewidth} p{0.8\linewidth}}
\toprule
target spans                                                   & [3, 5, 'PP'] \\ \hline
\multirow{3}{*}{\begin{tabular}[c]{@{}l@{}}DLM-E2E\\ float64\\ Standard FT\end{tabular}} & START There is a \textcolor{red}{\{\}} one star restaurant The Cambridge Blue . \textbackslash{}n END \\ & START The Cambridge Blue restaurant has English food \textcolor{red}{with} an average customer rating . \textbackslash{}n END \\ & START The Golden Curry has English food , kid friendly , \textcolor{red}{\{\}} riverside and has a customer rating of 1 out of 5 . It is located \textcolor{blue}{near} Café Rouge . \textbackslash{}n END                \\ \hline

\multirow{3}{*}{\begin{tabular}[c]{@{}l@{}}DLM-E2E\\ Q0i.8f\\ Standard FT\end{tabular}}  &  START It is located \textcolor{blue}{near} the Yippee Noodle Bar and it is called Strada . \textbackslash{}n END\\ & START The Waterman is located \textcolor{red}{in} the city centre . It has a low customer rating , is more than £ 30 . It is not children friendly , and is not child - friendly . \textbackslash{}n END\\ & START Bibimbap House is located \textcolor{blue}{near} Clare Hall in the city centre . \textbackslash{}n END            \\ \hline

\multirow{3}{*}{\begin{tabular}[c]{@{}l@{}}DLM-E2E\\ Q0i.4f\\ Standard FT\end{tabular}}  & START Indian coffee shop \textcolor{blue}{near} Café Sicilia is not family friendly . \textbackslash{}n END\\ & START It is located \textcolor{blue}{in} the riverside is not family friendly . \textbackslash{}n END \\ & START They are located \textcolor{blue}{in} city centre . \textbackslash{}n END \\ \hline

\multirow{3}{*}{\begin{tabular}[c]{@{}l@{}}DLM-E2E\\ Ternary\\ Standard FT\end{tabular}}  & START Express by Holiday Inn \textcolor{red}{in} the city centre serves English food called The Rice Boat . It has a high customer rating and is child friendly . \textbackslash{}n END \\ & START located \textcolor{blue}{in} the riverside area . It serves Indian food . It has a customer rating of 1 out of 5 . \textbackslash{}n END \\ & START Aromi is a pub located \textcolor{red}{at} City centre . its customer rating is 5 out of UNK . \textbackslash{}n END                                                                                                 \\ \hline
\multirow{3}{*}{\begin{tabular}[c]{@{}l@{}}DLM-E2E\\ Binary\\ Standard FT\end{tabular}}  & START It is located \textcolor{blue}{near} the Rainbow Vegetarian Café and is near Clare Hall . \textbackslash{}n END \\ & START Golden Curry \textcolor{red}{\{\}} . It is family - friendly , serves French food and has a low customer rating . \textbackslash{}n END \\ & START It is located \textcolor{blue}{in} the city centre near the Express by Holiday Inn , . Its customer rating is 1 out of 5 . \textbackslash{}n END              \\ \hline  \hline
target spans                                                   & [3, 5, 'PP']  \\ \hline

\multirow{3}{*}{\begin{tabular}[c]{@{}l@{}}DLM-E2E\\ float64\\ LoRA FT\end{tabular}}   & START Midsummer House \textcolor{blue}{near} Café Rouge is a fast food restaurant . \textbackslash{}n END \\ & START The Vaults , \textcolor{blue}{near} Café Adriatic , has a high price range . \textbackslash{}n END \\ & START The Plough is a cheap pub \textcolor{red}{near} Café Rouge that serves sushi and is family friendly . \textbackslash{}n END               \\ \hline

\multirow{3}{*}{\begin{tabular}[c]{@{}l@{}}DLM-E2E\\ Q0i.8f\\ LoRA FT\end{tabular}}    & START It is located \textcolor{blue}{in} near city centre and is not family friendly . \textbackslash{}n END \\ & START The Phoenix is located \textcolor{blue}{in} the city centre , and serves Indian food . It has an average customer rating . \textbackslash{}n END \\ & START It is located \textcolor{blue}{in} the riverside . It is near Express by Holiday Inn . Its customer rating is 1 out of 5 . \textbackslash{}n END              \\ \hline

\multirow{3}{*}{\begin{tabular}[c]{@{}l@{}}DLM-E2E\\ Q0i.4f\\ LoRA FT\end{tabular}}    & START It is located \textcolor{blue}{near} the river . It is \textcolor{blue}{near} The Rice Boat . \textbackslash{}n END \\ & START Golden Curry located \textcolor{blue}{near} Café Rouge , , is family friendly , and but has a low customer rating . \textbackslash{}n END \\ & START Golden Curry is located \textcolor{blue}{near} Café Rouge . It has a low price range and is not family friendly . \textbackslash{}n END  \\ \hline

\multirow{3}{*}{\begin{tabular}[c]{@{}l@{}}DLM-E2E\\ Ternary\\ LoRA FT\end{tabular}}    & START The Plough is \textcolor{blue}{near} Café Rouge . The UNK is UNK for UNK . \textbackslash{}n END\\ & START The Rice Boat , \textcolor{blue}{near} the river , is an English venue that is family friendly . \textbackslash{}n END \\ & START It is located \textcolor{blue}{in} the city centre and is not kid friendly . \textbackslash{}n END                                                                                              \\ \hline
\multirow{3}{*}{\begin{tabular}[c]{@{}l@{}}DLM-E2E\\ Binary\\ LoRA FT\end{tabular}}     & START It is located \textcolor{blue}{in} the riverside and is moderately priced . \textbackslash{}n END \\ & START It is family friendly and \textcolor{red}{\{\}} has a low customer rating . \textbackslash{}n END \\ & START Cambridge is a restaurant \textcolor{blue}{in} city centre near The Sorrento . It is family friendly . \textbackslash{}n END                 \\ \bottomrule
\end{tabular}
\end{table*}

\begin{table*}[ht]
\caption{Qualitative output from the semantic content control task is visually represented. Compliant spans are highlighted in green, while spans that deviate from the control target are marked in \textcolor{red}{red}. Words that match the target are distinguished by a \textcolor{blue}{blue} highlight.}
\label{Table3-16}
\scriptsize
\centering
\begin{tabular}{p{0.1\linewidth} p{0.8\linewidth}}
\toprule
\begin{tabular}[c]{@{}l@{}}target\\ semantic\\ content\end{tabular}                                                   & name : 'The', 'Vaults'    \\ \hline                                      

\multirow{3}{*}{\begin{tabular}[c]{@{}l@{}}DLM-E2E\\ float64\\ Standard FT\end{tabular}} & START The UNK for , \textcolor{blue}{The Vaults} restaurant UNK UNK is high . \textbackslash{}n END \\ & START \textcolor{blue}{The Vaults} is a high priced coffee shop with a 3 out of 5 rating located near Café Brazil by the riverside . \textbackslash{}n END \\ & START \textcolor{blue}{The Vaults} serves Chinese food in the high price range . \textbackslash{}n END                \\ \hline

\multirow{3}{*}{\begin{tabular}[c]{@{}l@{}}DLM-E2E\\ Q0i.8f\\ Standard FT\end{tabular}}  & START just \textcolor{blue}{Vaults} delicious a moderately priced coffee shop in also riverside near Café Brazil . \textbackslash{}n END\\ & START There delicious a Kid friendly coffee shop just \textcolor{blue}{Vaults} . It also a moderate price range range near Café Brazil in the riverside area boasts a customer rating of 1 out of 5 . \textbackslash{}n END\\ & START \textcolor{red}{Loch Fyne} delicious a your - friendly your restaurant with a price range less than £ 20 . \textbackslash{}n END             \\ \hline

\multirow{3}{*}{\begin{tabular}[c]{@{}l@{}}DLM-E2E\\ Q0i.4f\\ Standard FT\end{tabular}}  & START \textcolor{blue}{The Vaults} is located near Café Adriatic . The pub has a high customer rating . It is relatively expensive . \textbackslash{}n END\\ & START The Golden Curry is an Indian restaurant with a customer rating of 5 out of 5 . It is not family friendly . \textbackslash{}n END \\ & START \textcolor{blue}{The Vaults} is a moderately priced pub . It is located near Café Brazil . \textbackslash{}n END \\ \hline

\multirow{3}{*}{\begin{tabular}[c]{@{}l@{}}DLM-E2E\\ Ternary\\ Standard FT\end{tabular}}  & START The Eagle serves Indian food . It has a high customer rating . \textbackslash{}n END\\ & START \textcolor{blue}{The Vaults} is a family friendly fast food restaurant located in the city centre . \textbackslash{}n END \\ & START \textcolor{blue}{The Vaults} , a high priced pub , is located near Café Adriatic . \textbackslash{}n END                                                                                                 \\ \hline
\multirow{3}{*}{\begin{tabular}[c]{@{}l@{}}DLM-E2E\\ Binary\\ Standard FT\end{tabular}}  & START \textcolor{blue}{The Vaults} is a cheap restaurant that is family friendly . It has an average customer rating . \textbackslash{}n END\\ & START \textcolor{blue}{The Vaults} is a coffee shop with a high price . It is in high price range . It is located near', 'of 5 . It is not family friendly . \textbackslash{}n END \\ & START \textcolor{blue}{The Vaults} is a moderately priced , pub that is located near Café Brazil . \textbackslash{}n END                \\ \hline  \hline

target semantic content                                                   & name : 'The', 'Vaults'    \\ \hline

\multirow{3}{*}{\begin{tabular}[c]{@{}l@{}}DLM-E2E\\ float64\\ LoRA FT\end{tabular}}   & START \textcolor{blue}{The Vaults} , a family - friendly restaurant , serves Italian food with prices less than £ 20 . \textbackslash{}n END \\ & START \textcolor{blue}{The Vaults} is a family friendly place that serves low priced English food . ', 'range . \textbackslash{}n END \\ & START \textcolor{blue}{The Vaults} is great and offers traditional English food . \textbackslash{}n END               \\ \hline

\multirow{3}{*}{\begin{tabular}[c]{@{}l@{}}DLM-E2E\\ Q0i.8f\\ LoRA FT\end{tabular}}    & START \textcolor{blue}{The Vaults} is a coffee shop that is more moderately priced near Café Brazil . \textbackslash{}n END\\ & START \textcolor{blue}{The Vaults} is a not , family friendly restaurant ,', 'It is located in the riverside . \textbackslash{}n END\\ & START \textcolor{blue}{The Vaults} is a restaurant that serves Indian food ,', 'range , and is located in the riverside area . \textbackslash{}n END\             \\ \hline

\multirow{3}{*}{\begin{tabular}[c]{@{}l@{}}DLM-E2E\\ Q0i.4f\\ LoRA FT\end{tabular}}    & START The pub , \textcolor{blue}{The Vaults} , is cheap and located near Café Adriatic .\textbackslash{}n END\\ &  START \textcolor{blue}{The Vaults} is a high - priced , sushi restaurant , UNK , and is family friendly . \textbackslash{}n END \\ & START \textcolor{blue}{The Vaults} is a moderately priced pub . It is located near Café Adriatic . \textbackslash{}n END   \\ \hline

\multirow{3}{*}{\begin{tabular}[c]{@{}l@{}}DLM-E2E\\ Ternary\\ LoRA FT\end{tabular}}    & START \textcolor{blue}{The Vaults} is a cheap rated restaurant near Café Adriatic . \textbackslash{}n END\\ & START \textcolor{red}{The Golden Palace} is a coffee shop that is located in the city centre . It serves cheap fast food and has a customer rating of 5 out of 5 . \textbackslash{}n END \\ & START \textcolor{blue}{The Vaults} is a high priced ,', ', serves Italian food . It is not family - friendly . \textbackslash{}n END                                                                                              \\ \hline
\multirow{3}{*}{\begin{tabular}[c]{@{}l@{}}DLM-E2E\\ Binary\\ LoRA FT\end{tabular}}     & START \textcolor{blue}{The Vaults} is a', ', in the city centre area , it serves Indian food and is not family friendly . \textbackslash{}n END\\ & START \textcolor{blue}{The Vaults} is a expensive food restaurant with a three star . \textbackslash{}n END \\ & START \textcolor{red}{The Waterman} is a Chinese restaurant in the center of the city . It is not family friendly . \textbackslash{}n END                 \\ \bottomrule
\end{tabular}
\end{table*}

\end{document}